%% file: main.tex
\documentclass[runningheads]{llncs}

\usepackage{eccv}

\usepackage{eccvabbrv}

\usepackage{graphicx}
\usepackage{booktabs}

\usepackage[accsupp]{axessibility}  

\usepackage[pagebackref,breaklinks,colorlinks,citecolor=eccvblue]{hyperref}

\usepackage{orcidlink}

\begin{document}

\title{Geometry-Aware Single-Image 4D Synthesis via Dense Trajectory Generation} 

\titlerunning{MoGe4D}

\author{Yanran Zhang$^{*,1}$
\orcidlink{0009-0005-7209-9761}
\and
Ziyi Wang$^{*,1}$
\orcidlink{0000-0002-9007-1210}
\and
Wenzhao Zheng\textsuperscript{\Letter}$^{,1}$
\orcidlink{0000-0001-7188-3734}
\and
Zheng Zhu$^{2}$
\orcidlink{0000-0002-4435-1692}
\and
Jie Zhou$^{1}$
\orcidlink{0009-0001-7276-8487}
\and
Jiwen Lu$^{1}$
\orcidlink{0000-0002-6121-5529}}

\authorrunning{Zhang et al.}

\institute{$^1$ Department of Automation, Tsinghua University, China ~~~ 
$^2$ GigaAI\\
\url{https://ivg-yanranzhang.github.io/MoGe4D/}\\
\email{zhangyr21@mails.tsinghua.edu.cn, \ lujiwen@tsinghua.edu.cn}}
\newcommand{\ours}{{MoGe4D}\xspace}
\maketitle

\begin{figure}[h]
\vspace{-5mm}
\centering
\includegraphics[width=1\linewidth]{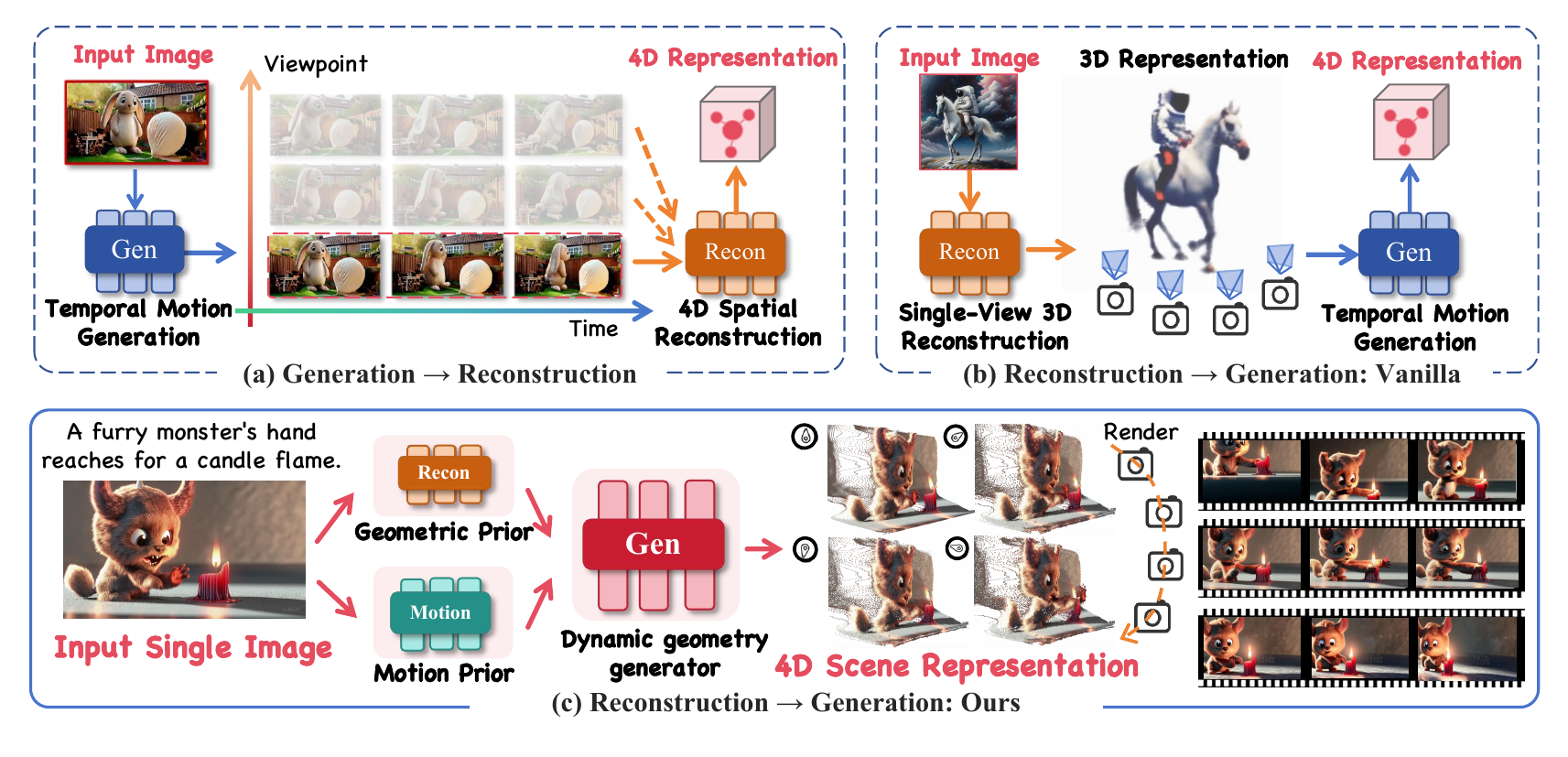}
\vspace{-8mm}
\captionof{figure}{\textbf{\ours for 4D synthesis from a single image.} Most existing paradigms either suffer from geometric inconsistencies (\textit{generate-then-reconstruct}) or are constrained by animating a pre-determined static geometry (vanilla \textit{reconstruct-then-generate}). Our \ours advances by tightly coupling geometric modeling and motion generation, effectively achieving consistent 4D motion and geometry.}
\vspace{-10mm}
\label{fig:teaser}
\end{figure}
\renewcommand{\thefootnote}{\fnsymbol{footnote}} 
\footnotetext[1]{Equal contributions. 
\textsuperscript{\Letter}Corresponding author.} 
\renewcommand{\thefootnote}{\arabic{footnote}}

\input{sec/0_abstract}

\input{sec/1_intro}

\input{sec/2_related_work}

\input{sec/3_method}
\input{sec/4_experiments}
\input{sec/5_conclusion}
\input{sec/6_acknowledgement}

{
\newpage
    \centering
    \Large
    \vspace{0.5em}Supplementary Material \\
    \vspace{1.0em}
}
\input{sec/supp}


%
%
\bibliographystyle{splncs04}
\bibliography{main}

\clearpage

\end{document}

%% file: sec/0_abstract.tex
\begin{abstract}
Generating interactive and dynamic 4D scenes from \textit{a single static image} remains a core challenge. Most existing \textit{generate-then-reconstruct} and \textit{reconstruct-then-generate} methods decouple geometry from motion, causing spatiotemporal inconsistencies and poor generalization. To address these, 
we present \textbf{\ours} (\textbf{Mo}tion and \textbf{Ge}ometry-Aware image-to-\textbf{4D} Synthesis), a geometry-conditioned framework for single-image 4D synthesis that models a scene as dense 4D point trajectories. Instead of treating geometry and dynamics as two disconnected stages, our method starts from an initial geometric prior inferred from the input image and predicts future time-varying trajectories in a diffusion process, improving spatiotemporal coherence while preserving structural stability.
To support this task, we first introduce TrajScene-60K, a large-scale dataset of 60,000 video samples with dense 4D point trajectories, addressing the scarcity of high-quality training data for scene-level 4D generation. Built on this, our diffusion-based 4D Scene Trajectory Generator (4D-STraG) predicts geometry-consistent and motion-plausible trajectory fields conditioned on the input image, with a depth-guided motion normalization strategy to reduce scale ambiguity and a Motion Perception Module (MPM) to inject motion-aware priors.
We further propose a 4D View Synthesis Module (4D-ViSM) to render the generated 4D representation into videos under arbitrary camera trajectories. Experiments show that \ours produces high-quality 4D scenes with strong temporal coherence, favorable geometry-aware consistency, and compelling novel-view synthesis from a single image. Code: \url{https://github.com/Zhangyr2022/MoGe4D}.

\keywords{4D Generation \and Dense Trajectories \and Geometry-Aware \and Novel View Synthesis \and Diffusion Models }
\end{abstract}

%% file: sec/1_intro.tex
\section{Introduction}
\label{sec:intro}

4D scene generation seeks to reconstruct comprehensive spatiotemporal representations capturing both explicit 3D geometry and complex temporal dynamics. Generating such dynamically rich 4D scenes from \textit{a single static image} remains a fundamental challenge~\cite{miao2025advances}, as the model must infer plausible future motion from a static observation while preserving structural coherence across time and viewpoints. Achieving this capability would largely benefit applications such as virtual reality, augmented reality~\cite{guo2023vid2avatar,miao2024pla4d}, and immersive content creation.

Although recent video generation models can produce realistic dynamic content, they generally lack an explicit understanding of 3D structure, leading to view inconsistencies and failing to capture physically plausible motion. Existing single-image-to-4D methods therefore typically adopt one of two decoupled paradigms. \textit{Generate-then-reconstruct} methods~\cite{cat4d,4real,dimensionX,free4d} use powerful video models to synthesize multi-view videos before reconstructing a 4D representation. This paradigm benefits from the high-fidelity and rich dynamics offered by existing generation models. 
However, video models struggle to maintain strict geometric consistency across the generated views, leading to significant artifacts and structural collapse during the subsequent 3D reconstruction phase. Consequently, the alternative paradigm, \textit{reconstruct-then-generate}~\cite{animate124,animate3d,gs-dit,jin2025optimizing, gen3c}, has emerged as another promising direction. By first establishing a static 3D structure, this approach provides a robust geometric foundation for subsequent motion generation. 
Still, by decoupling static geometry from motion generation, they discard the rich dynamic potential latent in the source image. As a result, they are restricted to modeling physically plausible and externally constrained motions (e.g., swinging) and struggle to generate large-scale and self-initiated movements that originate from the scene or objects themselves.

To address these challenges, we revisit the \textit{reconstruct-then-generate} paradigm and make its interface between structure and dynamics substantially tighter, as illustrated in Figure~\ref{fig:teaser}. Rather than explicitly refining the base geometry during generation, we condition future dynamic generation on an initial geometric prior inferred from the input image, and model the scene as dense 4D point trajectories, which is directly predicted from the single input image with an integrated model. To facilitate training of the 3D motion generation model, we first built TrajScene-60K, a large-scale dataset of 60,000 samples with 4D point trajectories, tackling data scarcity of large-scale, high-quality 4D scene data with complex dynamics. We then introduce the \textit{4D Scene Trajectory Generator} (4D-STraG) that takes an input image and predicts future time-varying point trajectories relative to the initial frame. Unlike prior works that treat these as separate steps, 4D-STraG operationalizes them within a unified denoising process, producing coherent 4D point trajectories whose motion is intrinsically consistent with the evolving 3D structure. To fully leverage priors from the input image, we further incorporate a depth-guided motion normalization method to enhance geometric awareness, and a Motion Perception Module (MPM) to leverage plausible motion priors. Finally, the generated 4D point cloud trajectories are rendered into high-fidelity dynamic videos from arbitrary novel viewpoints using our \textit{4D View Synthesis Module} (4D-ViSM), completing the full pipeline from a single image to a coherent 4D scene. 

Experimental results of both quantitative and qualitative comparisons demonstrate that our method consistently outperforms existing baselines, producing 4D scenes with more pronounced dynamics, stronger 3D motion consistency, and superior performance. Detailed ablation studies further confirm the effectiveness of our key modules, highlighting how each contributes to coherent and physically plausible 4D motion. Our contributions can be summarized as:

\begin{itemize}
    \item We construct TrajScene-60K, a large-scale 4D scene dataset featuring dense 4D point cloud trajectories, videos, and text to advance research in this field.
    \item We propose a geometry-conditioned dense-trajectory formulation for single-image 4D synthesis. It predicts scene trajectories from an initial geometric prior, avoiding the reliance on loosely decoupled stages.
    \item We develop 4D-STraG, a diffusion-based trajectory generator with depth-guided motion normalization and a Motion Perception Module for structural stability and motion plausibility, alongside 4D-ViSM for high-quality novel-view dynamic rendering from the generated 4D representation.
    \item Extensive experiments show that \ours delivers strong perceptual quality, geometry-aware consistency, and efficient 4D generation.
\end{itemize}

%% file: sec/2_related_work.tex
\section{Related Work}

\noindent\textbf{Video Generation.} The field aims to create temporally coherent and visually realistic dynamic visual content. From early VAE~\cite{vqvae,DBLP:conf/icml/DentonF18,he2018probabilistic,babaeizadeh2021fitvid,4dfy} and GAN~\cite{DBLP:conf/nips/VondrickPT16,tulyakov2018mocogan,clark2019adversarial} frameworks to modern large-scale diffusion models~\cite{VDM,make-a-video,khachatryan2023text2video,wan2025wan,kling} trained on extensive video data, the realism and resolution of generated videos have been revolutionized. In addition to text control generation, researchers have introduced various guidance strategies for video generation, such as structure-based~\cite{ma2024follow,xing2024make,chen2025echomimic}, image-based~\cite{chen2023videocrafter1,deng2024autoregressive}, and temporal controls~\cite{shi2024motion,wu2024motionbooth,yang2024direct}. However, most of them are fundamentally limited as they operate in 2D pixel space, lacking explicit 3D scene modeling. 

\vspace{4pt}
\noindent\textbf{Novel-View Synthesis (NVS).}
3D reconstruction-based methods reconstruct a 3D or 4D representation~\cite{4dgs,zhou2023nerflix,zhu2024fsgs,charatan2024pixelsplat}, ensuring strong geometric consistency but often requiring costly optimization and suffering from artifacts. Generation-based methods use pretrained video diffusion models conditioned on camera trajectories to synthesize novel views~\cite{Reconx,viewcrafter,cat3d,muller2024multidiff,bai2025recammaster}. While leveraging strong generative priors, they often exhibit inconsistencies and object drift under large viewpoint changes or complex scenes.

\vspace{4pt}
\noindent\textbf{4D Generation.}
4D generation aims to produce temporally evolving 3D representations from texts or sparse images. A growing body of work addresses 3D point tracking~\cite{karaev2024cotracker,karaev2025cotracker3,xiao2024spatialtracker, xiao2025spatialtrackerv2, delta, feng2025st4rtrack, lin2025dgs, lin2025movies} and video-to-4D reconstruction~\cite{xie2024sv4d, wang2025shape, sv4d2, wu2024sc4d, consistent4d, geo4d, yao2025uni4d,wu20254d,gs-dit,cat4d}, estimating dense spatiotemporal correspondences from multi-frame video inputs—a setting where rich temporal signals substantially constrain the problem. Image-to-4D is more ill-posed than video-to-4D, requiring full 3D and temporal reconstruction from single images without multi-frame priors. Existing Image-to-4D methods largely follow two decoupled paradigms: \textit{generate-then-reconstruct}, which first synthesizes videos and then reconstructs 4D (e.g., L4GM~\cite{l4gm}, 4Real~\cite{4real}, DimensionX~\cite{dimensionX}, Free4D~\cite{free4d}), and \textit{reconstruct-then-generate}, which builds static 3D assets before animating them (e.g., Animate124~\cite{animate124}, Animate3D~\cite{animate3d}, 4D-fy~\cite{4dfy}, Gen3C\cite{gen3c}). While effective, the former suffers from video–geometry mismatch, and the latter is often object-centric and limited in scene complexity. Recently, 4DNeX~\cite{4dnex} and One4D~\cite{mi2025one4d} concatenates RGB-XYZ, in contrast to our method, which advances the reconstruct-then-generate paradigm by modeling geometry-aware dense trajectory generation and enables consistent 4D synthesis.

%% file: sec/3_method.tex
\section{Dataset Curation -- TrajScene-60K}
\label{sec:data}
Developing a robust 4D generative framework necessitates a multi-faceted dataset comprising three essential modalities: dense 4D point trajectories, viewpoint-specific visual observations, and high-level semantic descriptions of 4D environments. To address the acute shortage of high-quality, large-scale annotated data---especially for complex scene-level videos featuring intricate motion and non-rigid dynamics---we present \textbf{TrajScene-60K}. This meticulously curated dataset provides the foundational supervision required for learning consistent 4D scene representations and trajectories.

\begin{figure*}[t]
\centering
\includegraphics[width=\textwidth]{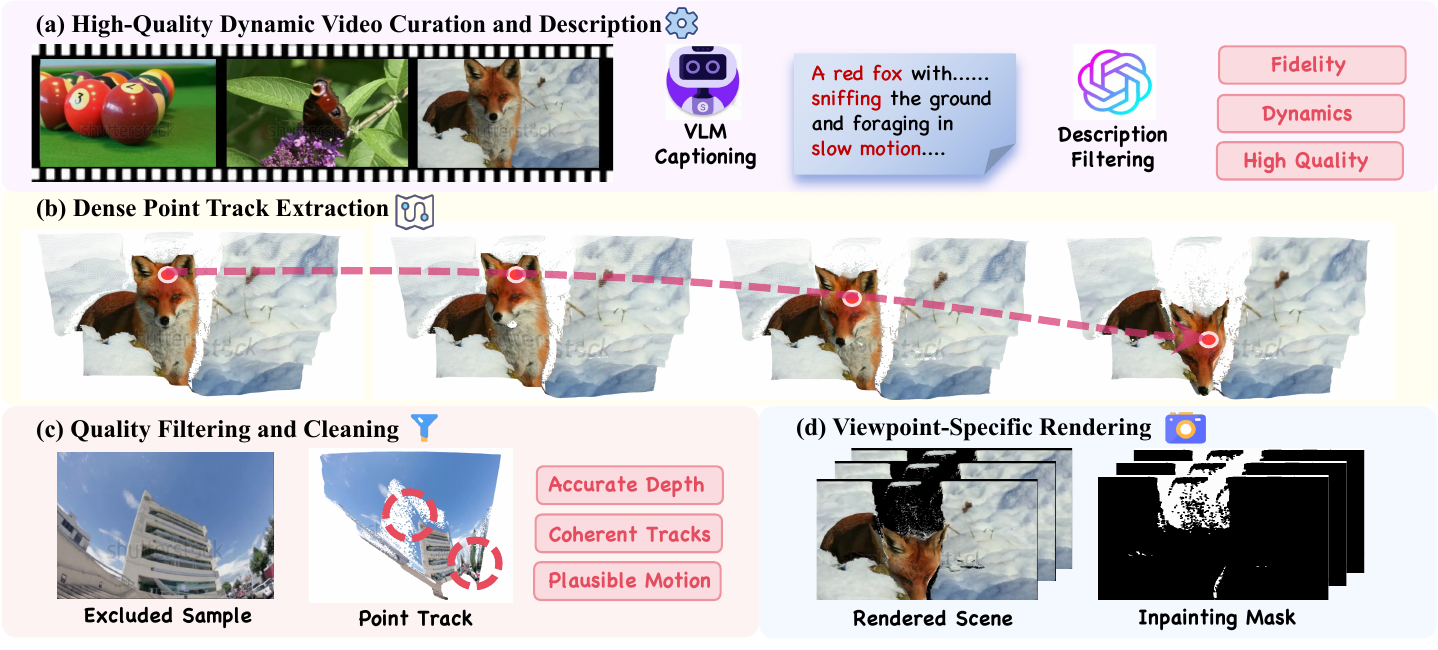}
\vspace{-6mm}
\caption{\textbf{TrajScene-60K curation pipeline.} We curate videos from WebVid-10M, filtered via VLMs for structured motion and countable entities. Dense 4D point tracks are extracted and refined via depth filtering and Gaussian Splatting, producing 60K high-quality 4D scenes.
}
\vspace{-8mm}
\label{fig:data}
\end{figure*}

As shown in Figure~\ref{fig:data}, we construct our dataset from the high-quality WebVid-10M corpus~\cite{webvid}, extracting about 200,000 candidates via a two-stage automated filtering pipeline. 
CogVLM2~\cite{cogvlm2} generates a detailed caption $\mathcal{C}$ for each video that captures both its visual content and the temporal characteristics of its motion.
And DeepSeek-V3~\cite{deepseekv3} evaluates these captions against two criteria: (1) the presence of one or more clearly countable entities, and (2) the exhibition of self-initiated, non-rigid, or articulated motion. 
Videos dominated by unstructured dynamics (e.g., crowd behaviors, wind-driven oscillations, or background jitter) or by camera motion are discarded. This is particularly important for 4D learning, since egomotion would otherwise be conflated with genuine object dynamics and introduce spurious global drift into the pseudo-GT trajectories. Retaining only clips whose motion originates from the scene itself thus yields well-structured, quantifiable motion targets that enable learning of clear and interpretable dynamic representations.

To extract 4D dense point track data from videos, we employ the DELTA~\cite{delta} model. The model takes RGB video sequences $\mathcal{V} \in \mathbb{R}^{T \times H \times W \times 3}$ as input and utilizes monocular depth estimation methods to obtain depth maps $\mathcal{D} \in \mathbb{R}^{T \times H \times W}$, subsequently estimating occlusion-aware 4D trajectories $\mathcal{P} \in \mathbb{R}^{T \times H \times W \times 4}$. Each 4D vector $\mathbf{p}_{t,u,v} = (u_t, v_t, d_t, o_t)$ represents the 3D position and occlusion status in frame $t$ corresponding to the pixel located at $(u, v)$ in the first frame. After obtaining 4D point cloud scene data, we apply a strict three-criterion quality filtering process: (1) samples containing trajectories with invalid or anomalous depth values (e.g., near-infinite or zero) are removed; (2) samples exhibiting excessively large standard deviation in scene depth, indicative of global depth estimation errors, are discarded; (3)
a scale consistency check removes geometrically inconsistent samples by verifying that uniformly scaled point clouds yield identical renderings from the original camera perspective.
This pipeline yields 60,000 high-quality samples at $596\times336$ resolution.

We then render these 4D scenes into videos from the original camera viewpoint using Gaussian Splatting~\cite{3dgs}, with inpainting masks generated for void regions. Compared to existing benchmarks, TrajScene-60K provides over \textbf{3 million frames} and approximately \textbf{12 billion 3D point annotations} with dense occlusion-aware tracking, per-frame depth, and language descriptions across diverse real-world indoor and outdoor environments---a combination of scale, realism, and multi-modal annotation unmatched by prior datasets. A detailed comparison with existing datasets is provided in the supplementary material.

\section{Proposed Approach}

\begin{figure*}[t]
\centering
\includegraphics[width=\textwidth]{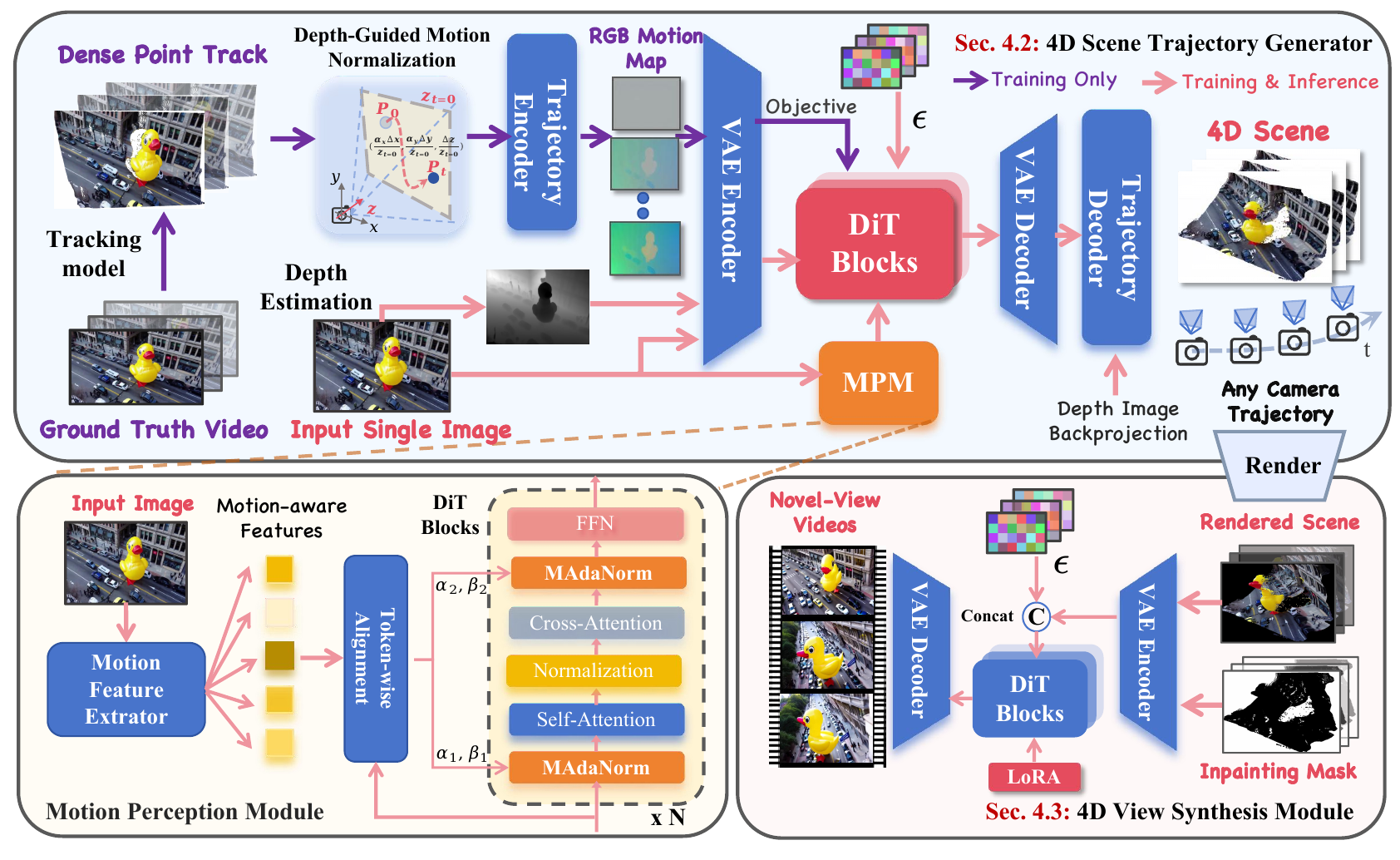}
\vspace{-6mm}

\caption{{\textbf{Pipeline of \ours.}} Top: The 4D Scene Trajectory Generator (Sec.~\ref{sec:4D-STraG}), a Diffusion Transformer, jointly generates future geometry and motion. Bottom-Left: The Motion Perception Module (MPM) identifies potential motion regions and semantic structure from the input image. Bottom-Right: The 4D View Synthesis Module (Sec.~\ref{sec:4D-ViSM}) renders the output into novel-view videos.}
\vspace{-2mm}
\label{fig:pipeline}
\end{figure*}

\subsection{Problem Definition}

Given a single input image $\mathcal{I} \in \mathbb{R}^{H \times W \times 3}$ and its description $\mathcal{C}$, our goal is to generate a physically plausible and temporally consistent 4D scene. We represent the scene as a point cloud sequence $\mathcal{P} \in \mathbb{R}^{T \times N \times 3}$, capturing the 3D geometry and motion trajectories of $N=H\times W$ points over $T$ frames. From this representation, we render dynamic scene videos $\mathcal{V} \in \mathbb{R}^{T \times H' \times W' \times 3}$ from arbitrary novel viewpoints or along arbitrary camera paths to the gap between a static image and full multi-view 4D dynamic generation.

Unlike canonical-mesh or template-based 4D representations, dense point trajectories are class-agnostic and require no predefined structure, decoupling geometry from appearance to flexibly accommodate diverse topologies and large deformations. This factorization provides a stable spatial reference and avoids the brittleness of loosely cascaded pipelines, while still allowing the model to synthesize rich scene dynamics in a unified trajectory space.
To systematically achieve these objectives, we first design \textbf{4D} \textbf{S}cene \textbf{Tra}jectory \textbf{G}enerator (\textbf{4D-STraG}), a diffusion model that predicts future 4D trajectory fields conditioned on the initial geometry from the initial image. Then we present in detail the \textbf{4D} \textbf{Vi}ew \textbf{S}ynthesis \textbf{M}odule (\textbf{4D-ViSM}), which leverages the reconstructed 4D representation to enable high-quality video generation under arbitrary camera motion. The overview illustration is shown in Figure~\ref{fig:pipeline}.

\subsection{ \textbf{4D} \textbf{S}cene \textbf{Tra}jectory \textbf{G}enerator (4D-STraG)}
\label{sec:4D-STraG}
 
To leverage the rich motion priors and structural understanding capabilities, we build upon Wan2.1~\cite{wan2025wan} and finetune its VAE and DiT modules separately.

\vspace{3pt}
\noindent\textbf{Point Trajectory Initialization and Normalization.}
To enhance training stability and ensure compatibility with generative model scales, we only use the 4D-STraG to predict relative motion $\mathbf{\Delta P}_t = \mathbf{P}_t - \mathbf{P}_0=\{[\Delta{x}_t,\Delta{y}_t,\Delta{z}_t]\}$, where $t \in [0, T]$ and $\mathbf{P}_0$ denotes coordinates in the first frame. We further normalize $\mathbf{\Delta P}_t$ to avoid an unlimited data value range. Based on the observation that a small 3D movement in a nearby object causes a large displacement on the 2D image plane, whereas the same movement in a distant object appears minuscule, we propose a \textit{Depth-Guided Motion Normalization Strategy}. It normalizes the absolute motion of each point relative to its viewing frustum at the initial depth. By doing so, we transform raw motion into a scale-invariant representation, ensuring perceptual consistency across different distances and producing a more uniform data distribution to learn from.

Specifically, given focal lengths $f_x, f_y$ and image size $W \times H$, we define the scaling factors $\alpha_x =  f_x / W $ and $\alpha_y =  f_y / H  $. Geometrically, $z / \alpha_x $ and $  z/\alpha_y$ correspond to the width and height of the viewing frustum at depth $z$, respectively. We normalize motion quantities by the viewing frustum size at depth $z = \mathbf{P}^{(z)}_{0}$:
\begin{equation}
\small
\Delta \tilde{x}_t = \frac{\alpha_x\cdot \Delta x_t }{z }, \quad \Delta \tilde{y}_t = \frac{\alpha_y\cdot \Delta y_t }{z}, \quad \Delta \tilde{z}_t = \frac{\Delta z_t}{z}.
\label{eq:norm}
\end{equation}
This depth-dependent normalization achieves scale invariance across different depth ranges, enabling our diffusion model to effectively learn motion patterns without being biased by the absolute spatial position of points. During inference, we use UniDepthv2~\cite{unidepthv2} to estimate depth, ensuring consistency with the DELTA tracking model setup. The relative motion maps are de-normalized and fused with the initial point cloud to form a 4D scene representation.

\vspace{3pt}
\noindent\textbf{Model Pipeline.}
During training, our model takes an image–caption pair as input and learns a diffusion model to predict pixel-level point trajectories across frames. To adapt the generative backbone, we first finetune a motion-sensitive VAE capable of handling trajectory signals. Specifically, the relative point displacement $\mathbf{\Delta P_t}$ is transformed into an RGB motion map by a lightweight Trajectory Encoder, where spatial movements are represented as color variations while the shape and appearance remain static as the first frame. Correspondingly, a Trajectory Decoder is appended after the VAE decoder, ensuring accurate recovery of point trajectories from the RGB motion map.

After finetuning the VAE, we adapt the Diffusion Transformer (DiT) to handle latents encoded from the RGB motion map. We explicitly inject strong geometric priors into the model, thereby enhancing its generative capability. Specifically, the depth information from the initial frame is encoded into a latent representation using the VAE encoder. This provides the model with robust structural priors and geometric cues, significantly improving its ability to reason about scene layout and object relationships. The image, noise, and depth latents are concatenated along the feature dimension to form the final input for the DiT: 
\begin{equation}
z_{\text{combined}} = \text{Concat}(z_{\text{image}}, z_{\text{noise}}, z_{\text{depth}}).
\label{eq:latent_concat}
\end{equation}

The DiT model is trained using flow matching~\cite{flow_matching}, which learns deterministic flows from noise to data distributions by minimizing the error between predicted and true flow fields, enabling accurate modeling of pixel-level motion. The objective function is defined as:
\begin{equation}
\mathcal{L}_{\mathrm{fm}} = \mathbb{E}_{t, x_0, x_1} \left[ | v_\theta(t, x_t) - (x_1 - x_0) |^2 \right],
\end{equation}
where $x_0$ and $x_1$ represent the initial and target states, respectively, $x_t$ denotes the interpolated state at time $t$, and $v_\theta$ is the flow vector predicted by the model.

\vspace{3pt}
\noindent\textbf{Motion Perception Module (MPM).} 
Adapting powerful motion priors from image-to-video diffusion models poses a key challenge. These models excel at generating dynamic, moving scenes from a static image. Our objective, however, is to leverage this pretrained temporal knowledge for a distinct task: generating structurally static videos that exhibit dynamic color variations. Here, the ‘dynamic color’ sequence is an intermediate RGB motion-map representation that encodes per-pixel 3D displacements. This creates a core conflict, as the model must learn to translate its ingrained understanding of physical displacement into a new domain of temporal color evolution.

To resolve this conflict and effectively guide the model's adaptation, we introduce the Motion Perception Module (MPM). The MPM is designed to identify semantic regions within the static scene that are plausible candidates for these color dynamics. We first employ a pretrained motion feature extractor, OmniMAE~\cite{omnimae}, to derive motion-aware patch-level features $\mathbf{S}$ from the input static image. To embed motion information into the diffusion process, we introduce \textit{Motion-aware Adaptive Normalization} (MAdaNorm). While inspired by conditional injection mechanisms like AdaLN~\cite{dit}, our proposed module is specifically designed for 4D motion generation. Unlike global conditioning approaches, MAdaNorm performs token-wise modulation driven by motion-aware features, enabling fine-grained control over temporal coherence in a geometrically consistent manner. Specifically, after spatial alignment of motion features $\mathbf{S}$ by resizing them to match the DiT token sequence length, token-wise adaptive parameters are generated through linear layers. For intermediate features $\mathbf{F}^i_t \in \mathbb{R}^{N \times d}$ in the $i$-th DiT block, the process is as follows:
\begin{gather}
\boldsymbol{\alpha}_1, \boldsymbol{\alpha}_2, \boldsymbol{\beta}_1, \boldsymbol{\beta}_2 = \text{Linear}(\mathbf{S}),\\
\mathbf{F}' = \text{Attn}\left( \boldsymbol{\gamma_1}\boldsymbol{\alpha}_1 \odot \text{LN}(\mathbf{F}^i_t) + \boldsymbol{\gamma_1}\boldsymbol{\beta}_1 \right), \\
\mathbf{F}'' = \text{MLP}\left( \boldsymbol{\gamma_2}\boldsymbol{\alpha}_2 \odot \text{LN}(\mathbf{F}') + \boldsymbol{\gamma_2} \boldsymbol{\beta}_2 \right),
\end{gather}
where $\boldsymbol{\alpha}_1, \boldsymbol{\alpha}_2, \boldsymbol{\beta}_1, \boldsymbol{\beta}_2 \in \mathbb{R}^{N \times d}$ are token-wise scaling and bias parameters, $\boldsymbol{\gamma_1},\boldsymbol{\gamma_2} \in \mathbb{R}^{d}$ are learnable global gating coefficient, and $\odot$ denotes token-wise multiplication. Experiments show that MPM improves the fidelity of motion trajectory modeling and enhances spatio-temporal consistency of generated 4D content.

\subsection{\textbf{4D} \textbf{Vi}ew \textbf{S}ynthesis \textbf{M}odule (\textbf{4D-ViSM})}
\label{sec:4D-ViSM}

After obtaining a dense 4D point cloud representation, we propose the 4D-ViSM to achieve novel view video synthesis along arbitrary camera trajectories. The original point cloud may not fully cover image regions in novel views, which results in holes in the rendered output.
Concretely, we rasterize each frame of the time-varying point cloud as a per-frame set of 3D Gaussians; temporally linking these per-frame Gaussians yields a lightweight dynamic renderer that avoids optimizing a full dynamic-4DGS field while substantially reducing projection-induced holes.
We thereby leverage generative models to complete these missing regions, ensuring visual coherence and plausibility.

Considering the excellent performance of Wan2.1~\cite{wan2025wan} in video generation tasks, we also choose to finetune this model to build our 4D-ViSM. The training data also come from our  TrajScene-60K dataset, including rendered videos, corresponding occlusion masks, ground truth videos, and captions. During training, we follow the Wan2.1 mask processing strategy, setting the mask value to 0.5 for regions without projected points in each frame. Through finetuning, our model achieves high-quality and visually consistent novel view video synthesis. 

\subsection{Inference}
\label{sec:inference}
Given a single image and a text prompt, we first estimate the initial depth with UniDepthv2~\cite{unidepthv2}, matching the estimator used by the DELTA tracker during training. The VAE then encodes the image and depth into latents while MPM extracts motion features, after which the DiT generates and decodes relative motion latents. These are de-normalized by reversing our depth-guided motion normalization and fused with the initial point cloud to construct the 4D scene, which 4D-ViSM finally renders from arbitrary camera poses into a spatio-temporally consistent novel-view video.

%% file: sec/4_experiments.tex
\section{Experiments}

\subsection{Experimental Setup}

\begin{figure*}[t]
\centering
\includegraphics[width=1.0\textwidth]{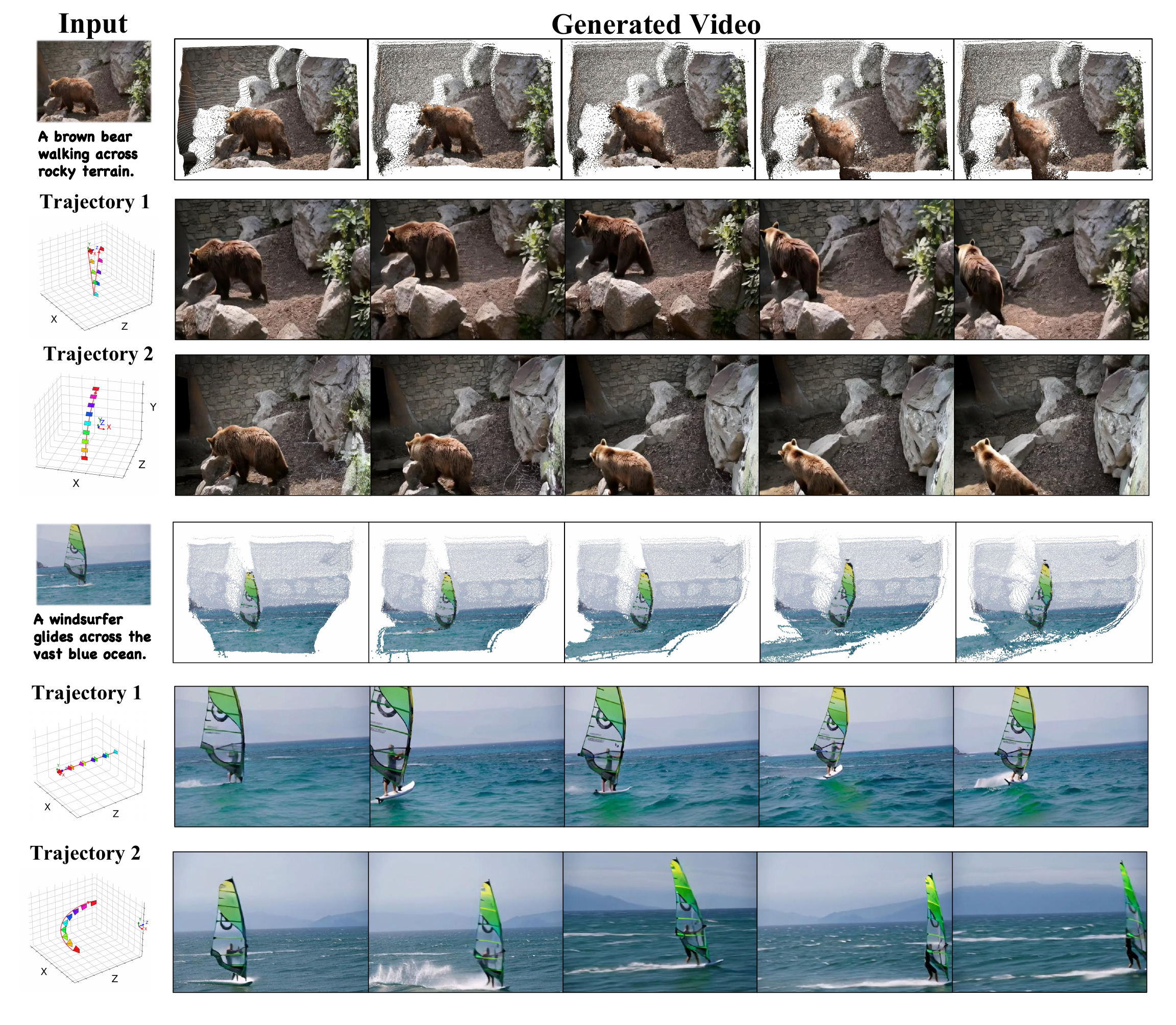}
\vspace{-15pt}
\caption{\textbf{Qualitative results of our model.} 
The first row shows the 4D point cloud generated by our 4D-STraG. The second and third rows show the videos rendered by our 4D-ViSM under two distinct, user-defined camera trajectories. }
\vspace{-5pt}
\label{fig:demo_main}
\end{figure*}

\noindent\textbf{Implementation Details.} 
Trained on TrajScene-60K dataset, our model generates  $512\times368$ videos with a length of 49 frames. For the 4D-STraG module, we performed full-parameter training based on Wan2.1-14B~\cite{wan2025wan,wan_fun}. The trajectory encoder and decoder are both shallow ResNets. We first trained its tracking components and VAE decoder for 5k steps, then its DiT for 2k steps using OmniMAE features for motion conditioning. The 4D-ViSM module finetuned Wan2.1-14B~\cite{wan2025wan,wan_fun_inp} with LoRA for 10k steps. We used AdamW~\cite{adamw} with a learning rate of $2\times 10^{-5}$. All experiments ran on four NVIDIA H20 GPUs.

\subsection{Qualitative Results}
\label{sec:qualitative}

\begin{figure*}[t!]
\centering
\includegraphics[width=1.0\textwidth]{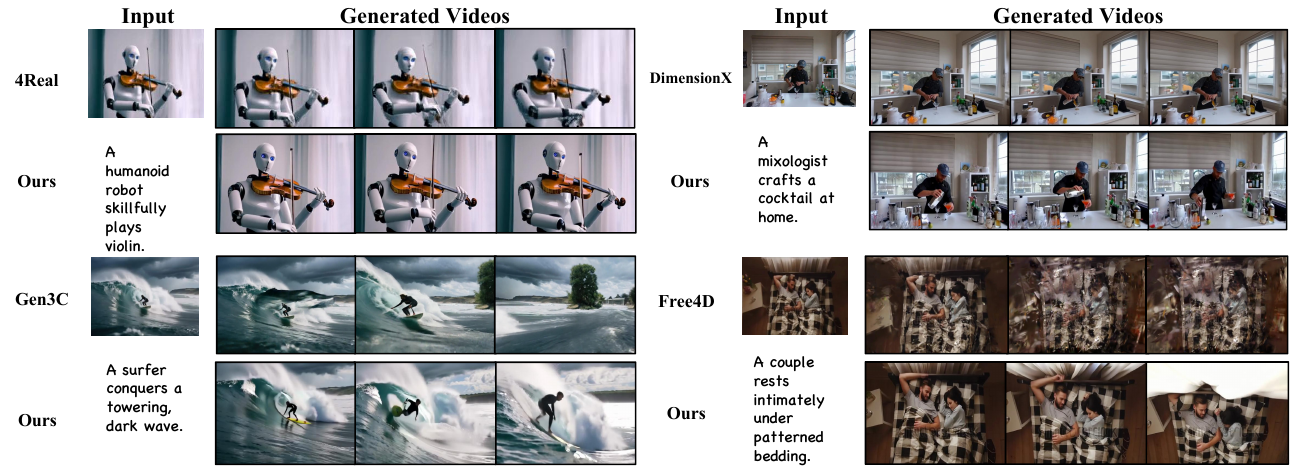}
\vspace{-16pt}
\caption{\textbf{Qualitative comparison with baseline methods.} For each sample, the first row shows the baseline results while the second row presents our \ours results. The first column displays the input image and text prompt.}
\label{fig:compare}
\vspace{-4pt}
\end{figure*}

For qualitative validation of our model effectiveness, we conduct comparative analyses in Figure~\ref{fig:demo_main} and Figure~\ref{fig:compare}. Figure~\ref{fig:demo_main} presents two representative scenes, with the first row displaying the point cloud outputs of 4D-STraG. It can be observed that the 4D point clouds maintain structural consistency and exhibit reasonable motion over time, with high detail completeness. For each scene, we define two camera trajectories and perform rendering using 4D-ViSM. The results demonstrate that the rendered images not only preserve the geometric accuracy of the original structure but also effectively align with the camera trajectories, reflecting strong visual consistency. Figure~\ref{fig:compare} further compares the visual outcomes of several existing methods, including 4Real, DimensionX, Gen3C, and Free4D. In comparison to these approaches, our method generates more diverse and realistic motion, validating the advantage of the our strategy in ensuring both structural rationality and motion plausibility for 4D generation.

\begin{table*}[t!]
\footnotesize  
\centering
\caption{\textbf{Quantitative comparison on VBench.} Higher values are better. The best results in each comparison group are marked in \textbf{bold}. Paradigm: Opt. = Optimization-based, Learn. = Learning-based, TF = Training-free.}
\vspace{-3mm}
\label{tab:vbench_results}
\renewrobustcmd{\bfseries}{\fontseries{b}\selectfont} 
\renewrobustcmd{\mdseries}{\fontseries{m}\selectfont}
\setlength{\tabcolsep}{6pt}
\small
\resizebox{\textwidth}{!}{%
\begin{tabular}{c|l|l|cccccc}
\toprule
\textbf{Exp.} & \multirow{2}{*}{\textbf{Model / Metrics}}
& \multirow{2}{*}{\textbf{Paradigm}}
& \textbf{Subject} & \textbf{Background}
& \textbf{Motion} & \textbf{Dynamic}
& \textbf{Aesthetic} & \textbf{Imaging} \\
 \textbf{No.} & & & {\textbf{Consistency}} & {\textbf{Consistency}}
& \textbf{Smoothness} & \textbf{Degree}
& \textbf{Quality} & \textbf{Quality} \\
\midrule
\multirow{2}{*}{\textbf{I}} & 4Real~\cite{4real} & Opt.   &  \bfseries  0.9329 & \bfseries  0.9709 & 0.9664 & 0.7708 & 0.4938 & 0.5095  \\
 & \ours(Ours) & Learn.    & 0.8752 &  0.9364      &    \bfseries 0.9682      &  \bfseries  1.0000       &   \bfseries  0.5613      &    \bfseries   0.6230    \\
\midrule
\multirow{3}{*}{\textbf{II}} & GenXD~\cite{genXD} & Learn.   & 0.8042 & 0.8789 & 0.9030 & \bfseries 1.0000 & 0.4077 & 0.5209 \\
 & DimensionX~\cite{dimensionX} & Learn. &  0.7553 &  0.8481 & \bfseries 0.9827 & \bfseries 1.0000 &  0.4634 &  0.5545 \\
 & \ours(Ours) & Learn.    & \bfseries 0.8241 & \bfseries 0.9044 &  0.9760 & 0.9500 & \bfseries 0.4820 & \bfseries 0.5828  \\
\midrule
\multirow{3}{*}{\textbf{III}} & Free4D~\cite{free4d} & TF     & 0.7899 & 0.8883 & 0.9797 & \bfseries 1.0000 & 0.3607 & 0.3562  \\
 & Gen3C~\cite{gen3c} & Learn.      & 0.8112 & 0.8871 & \bfseries  0.9845 & 0.9940 & 0.3812 & 0.4814  \\
 & \ours(Ours) & Learn.       & \bfseries 0.8339 & \bfseries 0.9065 & 0.9773 & 0.9000 & \bfseries 0.4820 & \bfseries 0.5939  \\
\bottomrule

\end{tabular}
}
\vspace{-4mm}
\end{table*}

\subsection{Quantitative Results}
\label{sec:quantitative}
\noindent\textbf{Video Quality Benchmarking.} Following prior 4D generation works~\cite{free4d}, we use VBench~\cite{vbench} for a fair and extensive video-quality comparison against a wider range of baselines. A more detailed 4D geometric analysis is provided in the supplementary material. We compare it with leading 4D generation methods: 4Real, GenXD, DimensionX, Gen3C, and Free4D. Following the evaluation protocol in Free4D~\cite{free4d}, we structured our comparisons into three groups based on model availability and technical constraints, with each group evaluated under appropriate camera trajectories. In Group I, we compared against the closed-source 4Real using simple trajectories, as we could only access their official demonstration videos for evaluation. Group II benchmarks single-image 3D models (GenXD, DimensionX's S-Director) with moderate 90° left rotations, due to limited trajectory support in their open-source code. For Group III, we evaluated against other 4D generation models using complex trajectories, including upward, forward, leftward, rightward, and downward 90° movements, to assess performance under challenging camera motions. For Groups II and III, we used 200 held-out samples from WebVid-10M that do not overlap with our TrajScene-60k training subset to ensure fair evaluation in diverse scenarios.  As shown in Table \ref{tab:vbench_results}, \ours consistently achieved promising results across groups. Notably, it demonstrated superior Dynamics, Aesthetic, and Imaging Quality compared to 4Real, outperformed 3D reconstruction baselines in consistency and visual quality, and maintained a pronounced lead in Aesthetic and Imaging Quality against 4D generation models under complex trajectories. Furthermore, consistent improvements across all groups demonstrate our robust generalization.

\begin{table*}[t!]\small
\small  
\centering
\caption{\textbf{Quantitative evaluation on 4D Generation Consistency via VLM-based assessment.} Higher values are better.}
\vspace{-3mm}
\label{tab:4d_video_results}
\sisetup{detect-weight, mode=text} 
\renewrobustcmd{\bfseries}{\fontseries{b}\selectfont} 
\renewrobustcmd{\mdseries}{\fontseries{m}\selectfont}
\setlength{\tabcolsep}{3pt} %
\resizebox{\textwidth}{!}{%
\begin{tabular}{c|l|ccccc|c}
\toprule
\textbf{Exp.} & \multirow{2}{*}{\textbf{Model / Metrics}} 
& \textbf{3D Geometric} & \textbf{Temporal Texture} 
& \textbf{Subject Identity} & \textbf{Motion Geometry} 
& \textbf{Background Stability} & \textbf{Average} \\
 \textbf{No.} & & {\textbf{Consistency}} & {\textbf{Stability}}  
& {\textbf{Preservation}} & {\textbf{Coupling}} 
& {\textbf{Consistency}} & \textbf{Score} \\
\midrule
\multirow{2}{*}{\textbf{I}} & 4Real~\cite{4real}   & 3.04 & 3.49 & 4.11 & 2.64 & 3.49 & 3.35 \\
 & \ours (Ours)    & \bfseries 3.46 & \bfseries 3.88 & \bfseries 4.42 & \bfseries 3.25 & \bfseries 3.96 & \bfseries 3.80 \\
\midrule
\multirow{3}{*}{\textbf{II}} & GenXD~\cite{genXD}   & 2.43 & 2.66 & 3.09 & 2.38 & 2.65 & 2.64 \\
 & DimensionX~\cite{dimensionX} & 2.58 & 2.74 & 3.35 & 2.53 & 2.79 & 2.80 \\
 & \ours (Ours)    & \bfseries 3.53 & \bfseries 3.92 & \bfseries 4.33 & \bfseries 3.45 & \bfseries 3.93 & \bfseries 3.83 \\
\midrule
\multirow{3}{*}{\textbf{III}} & Free4D~\cite{free4d}     & 1.13 & 1.30 & 1.37 & 1.17 & 1.17 & 1.23 \\
 & Gen3C~\cite{gen3c}      & 1.99 & 2.26 & 2.54 & 2.01 & 2.31 & 2.22 \\
 & \ours (Ours)       & \bfseries 3.47 & \bfseries 3.88 & \bfseries 4.25 & \bfseries 3.35 & \bfseries 3.85 & \bfseries 3.76 \\
\bottomrule
\end{tabular}
}
\vspace{-4mm}
\end{table*}

\noindent\textbf{4D Consistency Analysis.} Because standard metrics like VBench insufficiently penalize 3D-specific artifacts (e.g., geometric drift or texture swimming), we introduce a VLM-based protocol for comprehensive 4D evaluation. For videos generated across the aforementioned three groups, we uniformly sample 8 frames and prompt Qwen2.5-VL-72B-Instruct~\cite{bai2025qwen2} to rate them on a 1-5 scale across five criteria: 3D Geometric Consistency, Temporal Texture Stability, Subject Identity, Motion-Geometry Coupling, and Background Stability. We provide the exact scoring prompt and frame sampling strategy in the supplementary material for full reproducibility. As Table \ref{tab:4d_video_results} shows, \ours consistently outperforms all baselines. Notably, our significant margins in Motion-Geometry Coupling and Temporal Texture Stability validate \ours's superior spatial-temporal coherence and structural preservation during complex camera motions.

Moreover, to explicitly quantify structural stability, we conduct two rigorous geometry-aware evaluations: (1) \textbf{Avg. Trajectory Error}, measuring the $L_2$ distance between point trajectories tracked by DELTA in generated videos versus their ground-truth counterparts (evaluated on all generated samples from Groups II and III following DELTA's default tracking settings); and (2) \textbf{Avg. 3D Reprojection Error}, which follows WorldScore~\cite{zhang2025world} by utilizing DROID-SLAM~\cite{teed2021droid} to gauge the geometric deviation across all co-visible pixels in consecutive frames. As shown in Table~\ref{tab:3d_sub}, \ours achieves competitive 3D consistency. Notably, our method reduces the trajectory error by a substantial margin while maintaining highly competitive or superior reprojection errors. These results explicitly validate \ours's exceptional pixel-level tracking accuracy and robust rigid 3D spatial coherence.

\noindent\textbf{Inference Efficiency.} As Table~\ref{tab:runtime_comparison_sub} shows, unlike many optimization-based baselines that require hours of computation, \ours achieves highly competitive inference speed. On a single A100 GPU, it synthesizes 49-frame, $512\times368$ sequences in just 6 minutes. While GenXD is faster (2 mins), it yields significantly shorter, lower-resolution outputs. Crucially, we are $5\times$ faster than Free4D under equivalent resolutions. This remarkable efficiency allows our pipeline to inherently support longer-duration video synthesis.

\begin{table}[t]
\centering
\caption{\textbf{Runtime and error comparison.} Left: runtime on a single NVIDIA A100 (avg.\ 100 samples). Right: trajectory \& reprojection errors (lower is better)}
\label{tab:runtime_and_errors}
\setlength{\tabcolsep}{4pt}
\renewcommand{\arraystretch}{1.06}
  \vspace{-14pt} 
\begin{subtable}[t]{0.45\textwidth}
  \centering
  \caption{Mean Trajectory \& Reprojection Errors}
  \label{tab:errors_sub}
  \small
  \vspace{-10pt} 
  \resizebox{\linewidth}{!}{%
  \begin{tabular}{@{}c |l| c c@{}}
    \toprule
    \label{tab:3d_sub}
    \textbf{Exp.} & \textbf{Model} & \textbf{Trajectory Error} $\downarrow$ & \textbf{Reprojection Error} $\downarrow$ \\
    \midrule
    \multirow{4}{*}{\textbf{i}} 
      & \textcolor{gray}{Ground truth}           & \textcolor{gray}{0.000} & \textcolor{gray}{0.301} \\
      & GenXD~\cite{genXD}           & 0.236 & 1.891 \\
      & DimensionX~\cite{dimensionX} & 0.465 & \textbf{0.602} \\
      & \ours (Ours)                 & \textbf{0.058} & 0.614 \\
    \midrule
    \multirow{4}{*}{\textbf{ii}} 
      & \textcolor{gray}{Ground truth}           & \textcolor{gray}{0.000} & \textcolor{gray}{0.284} \\
      & Free4D~\cite{free4d}         & 0.252 & 0.652 \\
      & Gen3C~\cite{gen3c}           & 0.197 & 0.755 \\
      & \ours (Ours)                 & \textbf{0.042} & \textbf{0.639} \\
    \bottomrule
  \end{tabular}
  }
\end{subtable}
\hfill
\begin{subtable}[t]{0.48\textwidth}
  \centering
  \caption{Runtime Analysis(on a single A100)}
  \label{tab:runtime_comparison_sub}
  \small
  \vspace{-10pt} 
  \resizebox{\linewidth}{!}{%
  \begin{tabular}{@{}l| c c c c@{}}
    \toprule
    \textbf{Method} & \textbf{Year} & \textbf{Resolution} & \textbf{Frames} & \textbf{Time} \\
    \midrule
    4Dfy~\cite{4dfy}     & CVPR'24    & $256\times256$  & $-$   & 10 h   \\
    4Real~\cite{4real}   & NeurIPS'24 & $256\times144$  & 8     & 1.5 h  \\
    Gen3C~\cite{gen3c}   & CVPR'25    & $1280\times704$ & 121   & 50 min \\
    GenXD~\cite{genXD}   & ICLR'25    & $256\times256$  & 12    & 2 min  \\
    Free4D~\cite{free4d} & ICCV'25    & $512\times368$  & 16    & 30 min \\
    \midrule
    \ours (Ours)        & $-$        & $512\times368$  & 49    & 6 min  \\
    \bottomrule
  \end{tabular}
  }
\end{subtable}%

\vspace{-2pt}
\end{table}

\begin{table}[t!]
\centering
\caption{\textbf{Component-wise Ablation Study.} 
We analyze the impact of depth normalization, latent design, MPM structure, and training data variations. 
Best results per metric are highlighted in \textbf{bold}.}
\vspace{-4pt} 
\label{tab:ablation_transposed}
\footnotesize
\setlength{\tabcolsep}{4pt}
\renewcommand{\arraystretch}{1.05}
\resizebox{\textwidth}{!}{%
\begin{tabular}{l|c c c c c c c c}
\toprule
\multirow{2.5}{*}{\textbf{Metric}} &
\multirow{2.5}{*}{\makecell{\textbf{w/o Depth}\\\textbf{Norm}}} &
\multirow{2.5}{*}{\makecell{\textbf{w/o Depth}\\\textbf{Latents}}} &
\multicolumn{2}{c}{\textbf{MPM Module}} &
\multicolumn{3}{c}{\textbf{Training Data}} &
\multirow{2.5}{*}{\textbf{\ours}} \\
\cmidrule(lr){4-5} \cmidrule(lr){6-8}
& & & \textbf{w/o MPM} & \textbf{w/o Patch Feat.} 
& \textbf{Small Scale} & \textbf{Low Quality} & \textbf{Noise} & \\
\midrule
\textbf{Consistency↑} 
& 0.8604 & 0.8567 & 0.8650 & \textbf{0.8743} 
& 0.8567 & 0.8549 & 0.8685 & 0.8702 \\

\textbf{Dynamic↑}     
& 0.8850 & 0.8500 & 0.8500 & 0.8840 
& 0.8920 & 0.8850 & 0.8967 & \textbf{0.9000} \\

\textbf{Aesthetic↑}   
& 0.4672 & 0.4738 & 0.4806 & 0.4754 
& 0.4791 & 0.4654 & 0.4771 & \textbf{0.4820} \\
\bottomrule
\end{tabular}
}
\vspace{-4pt} 
\end{table}

\begin{figure*}[t]
\centering
\includegraphics[width=1.0\textwidth]{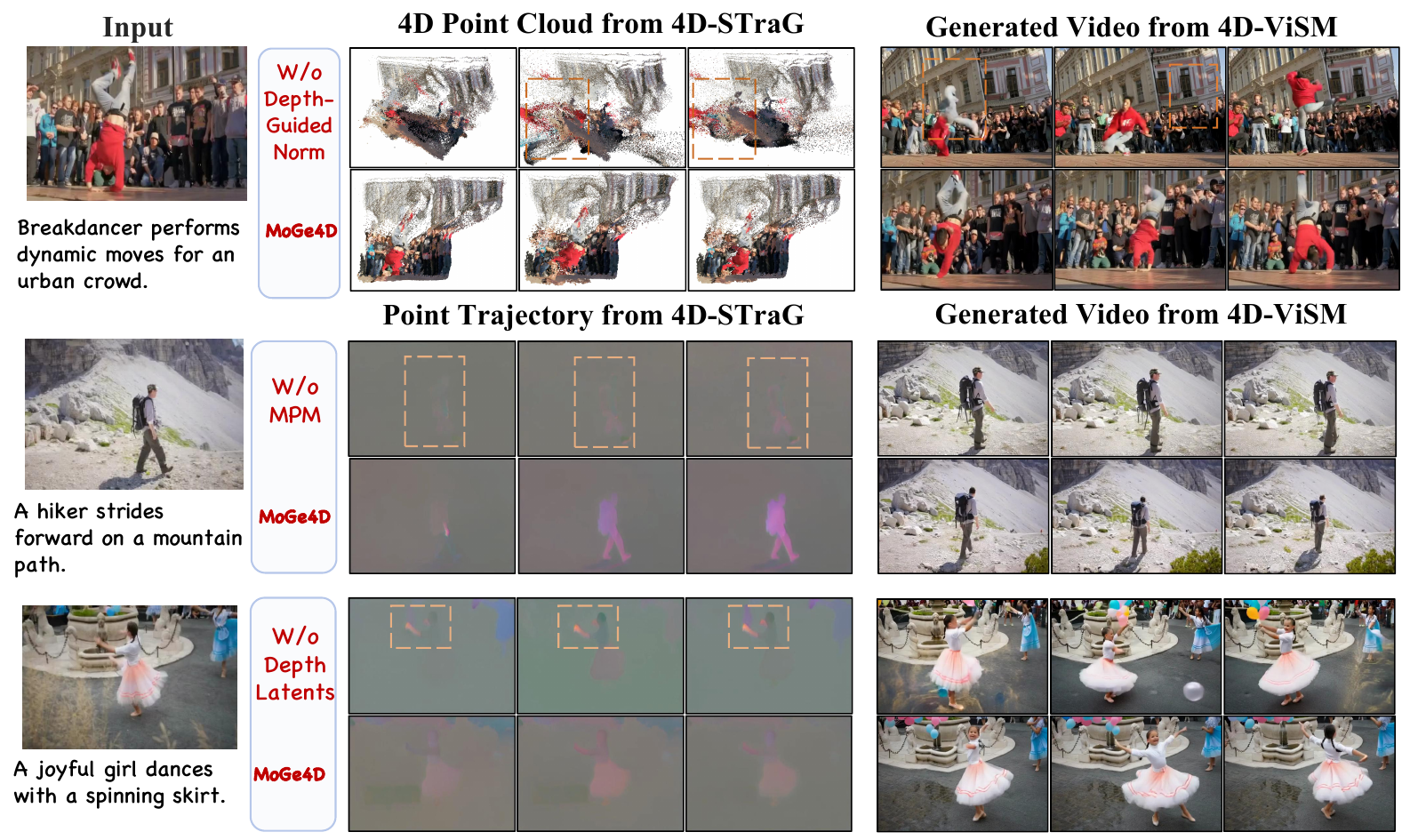}
\vspace{-10pt}
\caption{\textbf{Ablation studies on normalization methods and module components.} (Rows 1-2) Depth-guided motion normalization stabilizes 4D point cloud generation. (Rows 3-6) Removing the MPM module reduces motion magnitude while excluding depth guidance breaks structural motion consistency, validating our design choices.}
\vspace{-3pt}
\label{fig:ablation_module}
\end{figure*}

\begin{figure}[t]
  \centering
  \includegraphics[width=\linewidth]{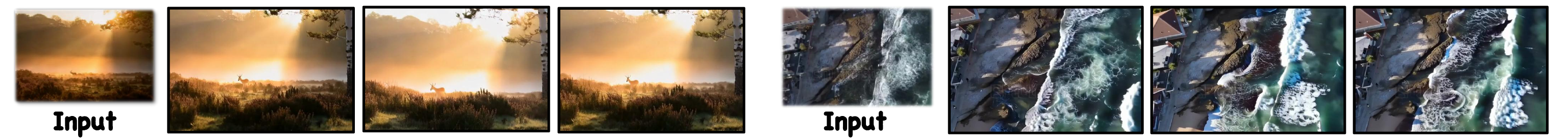}
  \vspace{-8pt}
  \caption{Limitations: Extreme lighting and unstructured crowds.}
   \label{fig:failure}
   \vspace{-12pt}
\end{figure}

\subsection{Ablation Studies}

Table~\ref{tab:ablation_transposed} presents a comprehensive ablation study on VBench under the Group III setting. Removing depth-guided normalization drops Consistency to 0.8604 and Aesthetic quality to 0.4672, while excluding depth latents degrades both Consistency and Dynamic scores, as the model struggles to maintain motion coherence. Removing MPM reduces Dynamic from 0.9 to 0.85; replacing its patch-level features with a global \texttt{[CLS]} token slightly improves Consistency but degrades Dynamic and Aesthetic quality, confirming that fine-grained patch-level conditioning is essential for high-fidelity 4D generation. For training data, reducing scale to a 1k subset or using unfiltered 60k samples both cause significant performance drops, validating that large-scale, high-quality data is foundational. Notably, adding random noise during training yields only negligible degradation, demonstrating robustness to noisy pseudo-GT supervision. Our full \ours achieves optimal performance across all metrics, validating the synergistic contribution of each component.

Row 1-6 in Figure~\ref{fig:ablation_module} visually demonstrate these findings. Baseline min-max normalization (rows 1-2) yields unstable point clouds with exaggerated movements, particularly in scenes with large depth variations. Our depth-guided approach effectively resolves this issue. Further visualizations (rows 3-6) show that without MPM or depth guidance, the model fails to preserve object structure or maintain motion consistency across different parts, confirming the critical role of depth in ensuring structurally coherent point trajectories.

\subsection{Discussions}

\begin{figure*}[t]
\centering
\includegraphics[width=1\textwidth]{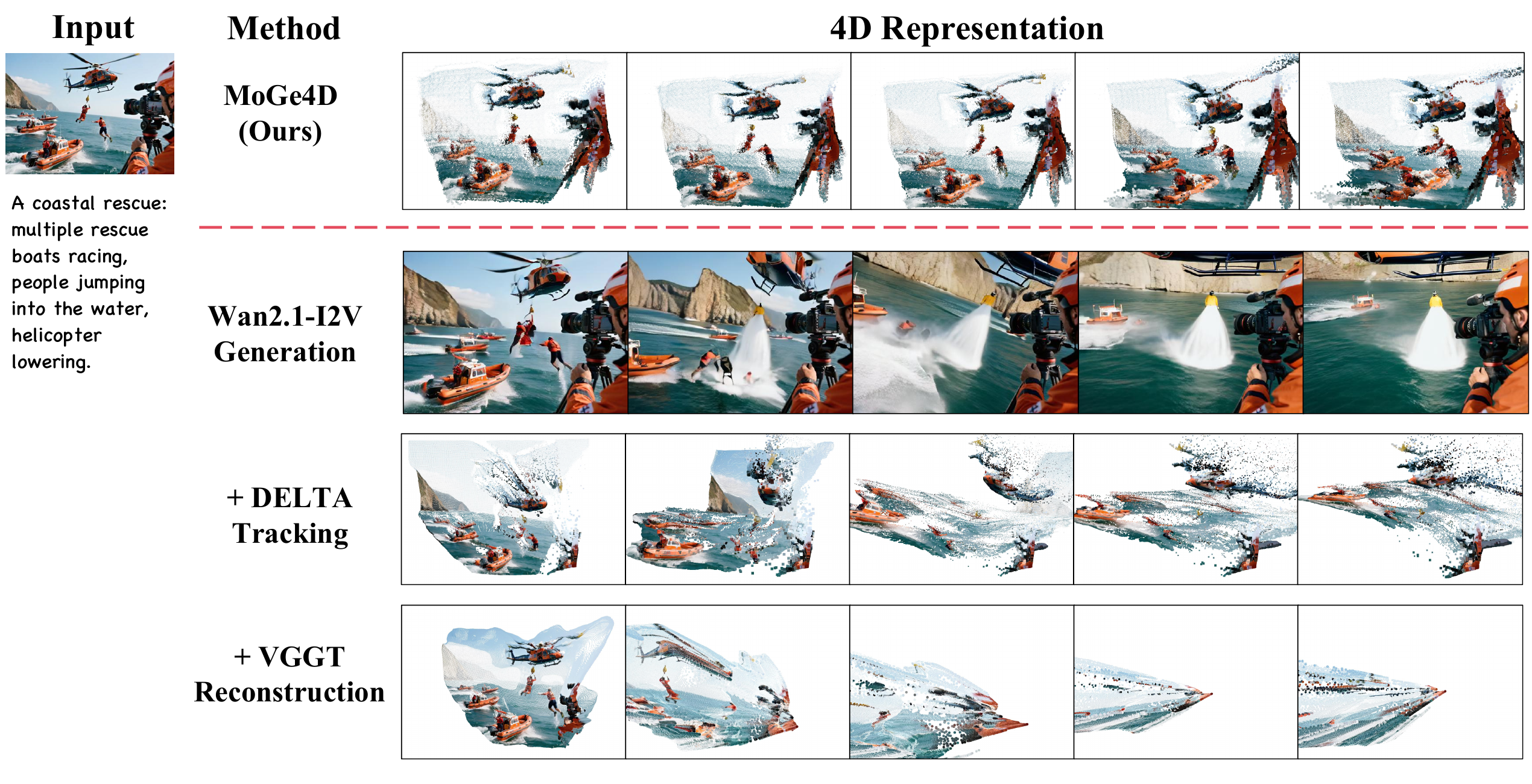}
\vspace{-10pt}
\caption{\textbf{Geometry-conditioned 4D-STraG (top row) vs.\ sequential pipelines.} Rows 2--4 show cascaded baselines: Wan2.1-I2V video generation followed by DELTA tracking or VGGT reconstruction. By conditioning trajectory generation directly on initial point cloud geometry, our method achieves superior spatio-temporal coherence, whereas sequential approaches suffer fragmentation and drift from error accumulation. All results are rendered from a fixed camera viewpoint for fair comparison.}
\vspace{-4pt}

\label{fig:vggt_comp}
\end{figure*}

\noindent\textbf{Geometry-Conditioned vs.\ Sequential 4D Generation.} To validate our geometry-conditioned trajectory generation paradigm, we compare against two sequential baselines: Wan2.1-I2V~\cite{wan2025wan} followed by either DELTA~\cite{delta} tracking or VGGT~\cite{vggt} reconstruction. As shown in Figure~\ref{fig:vggt_comp}, sequential pipelines suffer from severe error accumulation: since video diffusion models lack explicit 3D constraints, frame-level appearance inconsistencies are misinterpreted as 3D motion by downstream modules, producing spurious background drift and geometric fragmentation. In contrast, our approach conditions trajectory generation directly on the initial point cloud geometry, enabling the diffusion model to produce temporally coherent, geometry-consistent 4D trajectories without relying on error-prone intermediate video synthesis.

\noindent\textbf{Limitations.}
As shown in Fig.~\ref{fig:failure}, \ours struggles with extreme photometric conditions (e.g., harsh backlighting) and highly unstructured dynamics (e.g., chaotic crowds). In these complex edge cases, off-the-shelf monocular depth estimators often degrade, introducing noise into the initial geometric prior. Since our dense trajectory generation is conditionally bounded by the quality of this initial geometry, the resulting motion can become less reliable. Integrating more robust visual foundation models to handle such extreme physical conditions remains an important direction for future work.

%% file: sec/5_conclusion.tex
\section{Conclusion}

We proposed \ours, a geometry-conditioned framework for single-image 4D synthesis that represents a scene as dense time-varying point trajectories, overcoming weaknesses of decoupled paradigms. To support training, we constructed TrajScene-60K, a large-scale dataset with dense 4D trajectories, and introduced the 4D-STraG diffusion model with depth-guided motion normalization and motion perception priors. Experiments show that \ours achieves superior geometric consistency, dynamic realism, and visual fidelity compared to existing approaches. This work establishes a new technical pathway for 4D synthesis from minimal input, and points toward future exploration of more universal dynamic priors and lightweight 4D representations for practical deployment.

%% file: sec/6_acknowledgement.tex
\section{Acknowledgement}

This work was supported by the National Natural Science Foundation of China under Grant 62336004, Grant 623B2063, Grant 62125603, Grant 62576188, and Grant 62321005.

%% file: sec/supp.tex
\renewcommand\thesection{\Alph{section}} 
\renewcommand\thetable{\Alph{table}}
\renewcommand\thefigure{\Alph{figure}}
\setcounter{section}{0}
\setcounter{figure}{0}
\setcounter{table}{0}

\noindent In this Appendix, we provide:
\begin{itemize}
    \item Additional experimental settings and evaluation details (Section \ref{sec:a}).
    \item Additional discussions and comparisons (Section \ref{sec:b}).
    \item Extended visualization results (Section \ref{sec:c}).
    \item Dataset curation and comparison details (Section \ref{sec:d}).
    \item Baseline introductions (Section \ref{sec:e}).
    \item Implementation details (Section \ref{sec:f}).
    \item Applications and future directions (Section \ref{sec:g}).
    \item Prompt for VLM-based 4D evaluation (Section \ref{sec:prompt}).
\end{itemize}

\section{More Experiment Settings}
\label{sec:a}

\subsection{VBench Metrics}

\label{app:vbench_metrics}

In Table 1, we selected VBench~\cite{vbench} for quantitative evaluation as it offers a standardized and efficient framework to access perceptual video quality across diverse baselines, whereas 4D geometric metrics are computationally intensive and incompatible with many baseline methods. VBench comprehensively evaluates six critical dimensions:

\begin{itemize}
    \item {Subject Consistency}: Measures identity and appearance coherence of the primary subject over time.
    \item {Background Consistency}: Assesses temporal stability and coherence of background elements.
    \item {Motion Smoothness}: Evaluates the naturalness and fluidity of object and camera motion.
    \item {Dynamic Degree}: Quantifies the intensity and extent of dynamic changes in the scene.
    \item {Aesthetic Quality}: Judges overall visual appeal, composition, and stylistic merit.
    \item {Imaging Quality}: Rates technical image attributes like sharpness, noise, and artifacts.
\end{itemize}

This multi-dimensional analysis ensures a fair and holistic comparison of performance on 4D generation from a single image. 

\subsection{Details on VLM-based 4D Evaluation}

To address this evaluation gap, we leverage a Vision-Language Model (VLM) for systematic quality assessment in the main paper, Table 2. We uniformly sample 8 frames from each video generated under the settings of Table~1 in the main paper and feed them to Qwen2.5-VL-72B-Instruct~\cite{bai2025qwen2}, which rates the sequence on a 1-5 scale across five critical dimensions: (a detailed prompt is provided at the end section~\ref{sec:prompt})

\begin{itemize}
\item 3D Geometric Consistency: assesses 3D coherence during camera motion.
\item Temporal Texture Stability: evaluates texture stability without flickering.
\item Subject Identity Preservation: ensures subjects retain identity and shape.
\item Motion-Geometry Coupling: checks motion alignment with 3D geometry.
\item Background Stability: measures background static behavior.
\end{itemize}

\subsection{Details on 3D Geometry-Aware Evaluations}
\label{app:geometry_metrics}

While standard video evaluation metrics (such as VBench) provide comprehensive assessments of 2D perceptual quality and temporal smoothness, they often fail to penalize 3D-specific artifacts such as geometric structural collapse or ``texture swimming.'' To explicitly validate the 3D spatial coherence of our generated 4D scenes, we introduce two rigorous geometry-aware quantitative metrics.

\noindent\textbf{Average Trajectory Error.}
We employ the DELTA tracking model under its default settings to extract dense 3D point trajectories from both generated and ground-truth videos. The metric is defined as the average $L_2$ distance between predicted and ground-truth 3D point locations over all valid tracked frames:
\begin{equation}
    e_{\text{traj}} = \frac{1}{\sum_{t,i} \mathbf{M}_{i,t}} \sum_{t=1}^{T} \sum_{i=1}^{N} \mathbf{M}_{i,t} \left\| \hat{\mathbf{P}}_{i,t} - \mathbf{P}^*_{i,t} \right\|_2
\end{equation}
where $\hat{\mathbf{P}}_{i,t}$ and $\mathbf{P}^*_{i,t}$ are the 3D coordinates of point $i$ at frame $t$ in generated and ground-truth videos respectively, and $\mathbf{M}_{i,t} \in \{0, 1\}$ denotes point visibility. This metric strictly penalizes unnatural movements and spatial drift.

\noindent\textbf{3D Reprojection Error.}
To quantify rigid 3D structural stability, we adopt the dense reprojection error metric following WorldScore~\cite{duan2025worldscore} in Table 3(a). We utilize DROID-SLAM~\cite{teed2021droid}, which employs a differentiable Dense Bundle Adjustment (DBA) layer to continuously refine camera poses and per-pixel depth estimates. The reprojection error is then computed across all co-visible points between consecutive frames:
\begin{equation}
    e_{\text{reproj}} = \frac{1}{|\mathcal{V}|} \sum_{(i,j) \in \mathcal{V}} \left\| \mathbf{p}^*_{ij} - \Pi(\mathbf{P}_{ij}) \right\|_2^2
\end{equation}
where $\mathcal{V}$ is the set of co-visible pixel pairs, $\mathbf{p}^*_{ij}$ is the observed 2D coordinate in the target frame, $\mathbf{P}_{ij}$ is the reconstructed 3D point from refined depth and camera pose, and $\Pi(\cdot)$ denotes the camera projection. Since DROID-SLAM aligns all available pixels rather than relying on sparse features, a low $e_{\text{reproj}}$ rigorously confirms that the generated sequence maintains coherent 3D geometry across frames.

\begin{figure}[t!]
\centering
\includegraphics[width=1\textwidth]{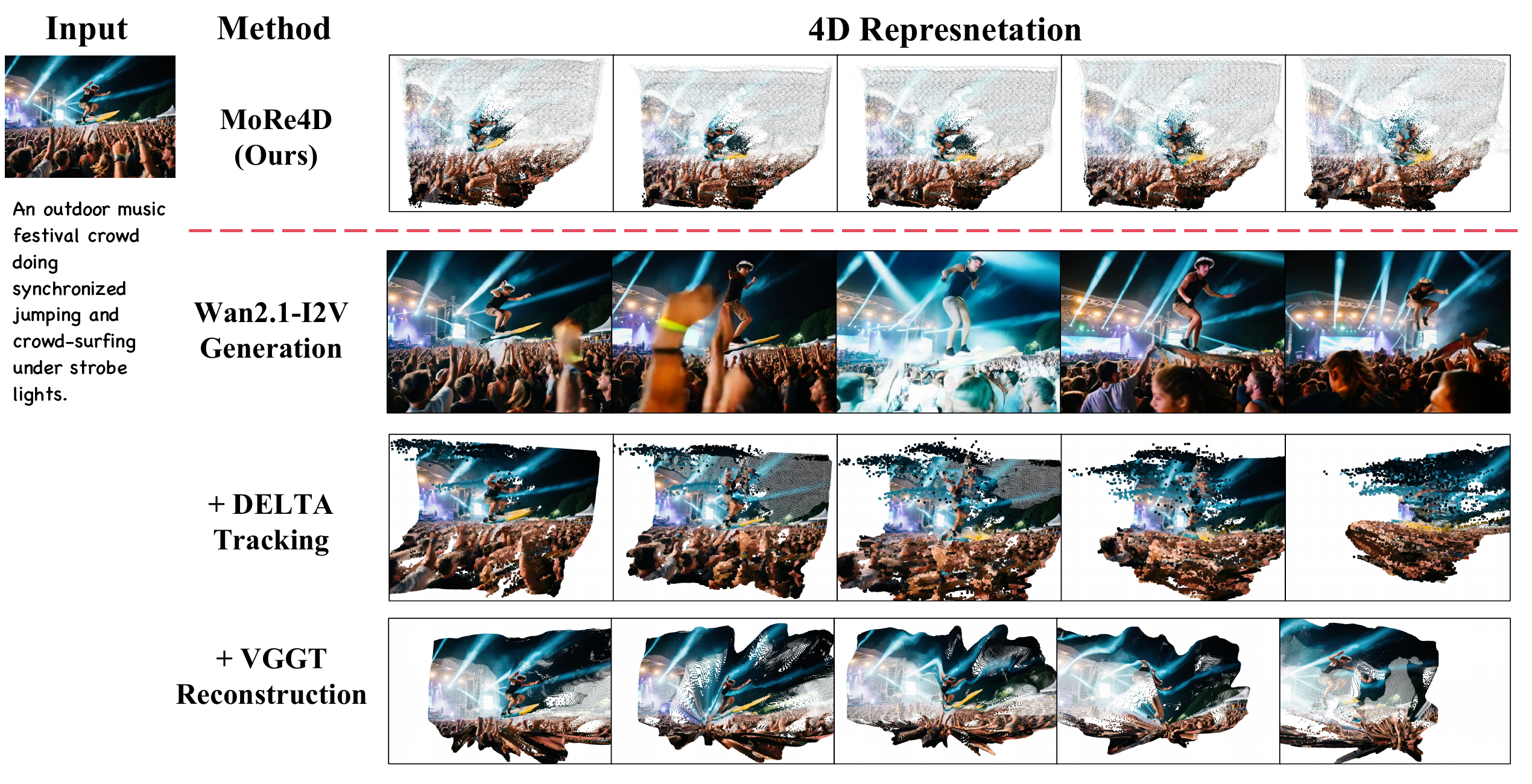}
\vspace{-20pt}
\caption{\textbf{An example comparing our geometry-conditioned 4D-STraG framework (top row) against sequential pipelines.} For each sample, rows 2-4 show results from a cascaded approach: Wan2.1-I2V video generation, followed by DELTA tracking or VGGT reconstruction. Our method yields superior spatio-temporal coherence, while sequential approaches exhibit fragmentation from error accumulation. All samples are consistently rendered from the fixed camera viewpoint for fair comparison.}
\label{fig:comp}
\vspace{-6pt}
\end{figure}

\section{More Discussions}
\label{sec:b}

\noindent\textbf{Geometry-Conditioned vs. Sequential 4D Generation.} To validate the effectiveness of our {geometry-conditioned trajectory generation paradigm}, we conduct an additional ablation study against alternative sequential pipelines that decompose the task into two stages: first generating a video from the source image, then performing 3D reconstruction or tracking as post-processing to obtain 4D representations. Specifically, we construct two baseline pipelines: (1) video generation using Wan2.1-I2V-14B~\cite{wan2025wan} followed by DELTA~\cite{delta}, the same dense point tracking model used in our dataset construction pipeline; and (2) the same video generation followed by VGGT~\cite{vggt}, a state-of-the-art feed-forward 3D transformer designed for robust 4D scene reconstruction.

As illustrated in Figure~\ref{fig:comp}, where results are visualized under a fixed rendering viewpoint to reveal geometric consistency, our geometry-aware framework demonstrates substantially superior spatio-temporal coherence. The generated point clouds maintain structurally coherent geometry and physically plausible motion throughout the sequence. In contrast, sequential pipelines exhibit notable failure modes: stationary background regions exhibit spurious motion (e.g., drifting water and crowds), while dynamic foreground objects suffer from geometric fragmentation and discontinuous trajectories (e.g., unrealistic body part detachment). These artifacts stem from spatial inconsistencies introduced during video generation—since standard video diffusion models like Wan-I2V lack explicit 3D geometric constraints, the synthesized videos contain frame-to-frame appearance variations, motion blur, and perspective inconsistencies that violate rigid scene structure. When fed into subsequent reconstruction or tracking modules, these visual inconsistencies are incorrectly interpreted as 3D motion or deformation, causing severe error accumulation.

Our method fundamentally mitigates these issues by conditioning dense trajectory prediction on the geometric prior within a unified diffusion backbone. This enables geometry-guided motion synthesis: the static structural constraints effectively regularize the denoising process toward spatial consistency, while visual features inform the trajectory prediction for temporal coherence. By learning to evolve scene dynamics directly from the base geometry, our approach ensures that every frame is both visually plausible and geometrically self-consistent under the fixed viewpoint. This prevents the error accumulation plaguing sequential approaches, ensuring robust and high-fidelity 4D content creation with coherent structure, realistic appearance, and accurate motion dynamics.

\section{More Visualization Results}
\label{sec:c}

\begin{figure*}[t!]
\centering
\includegraphics[width=0.95\textwidth]{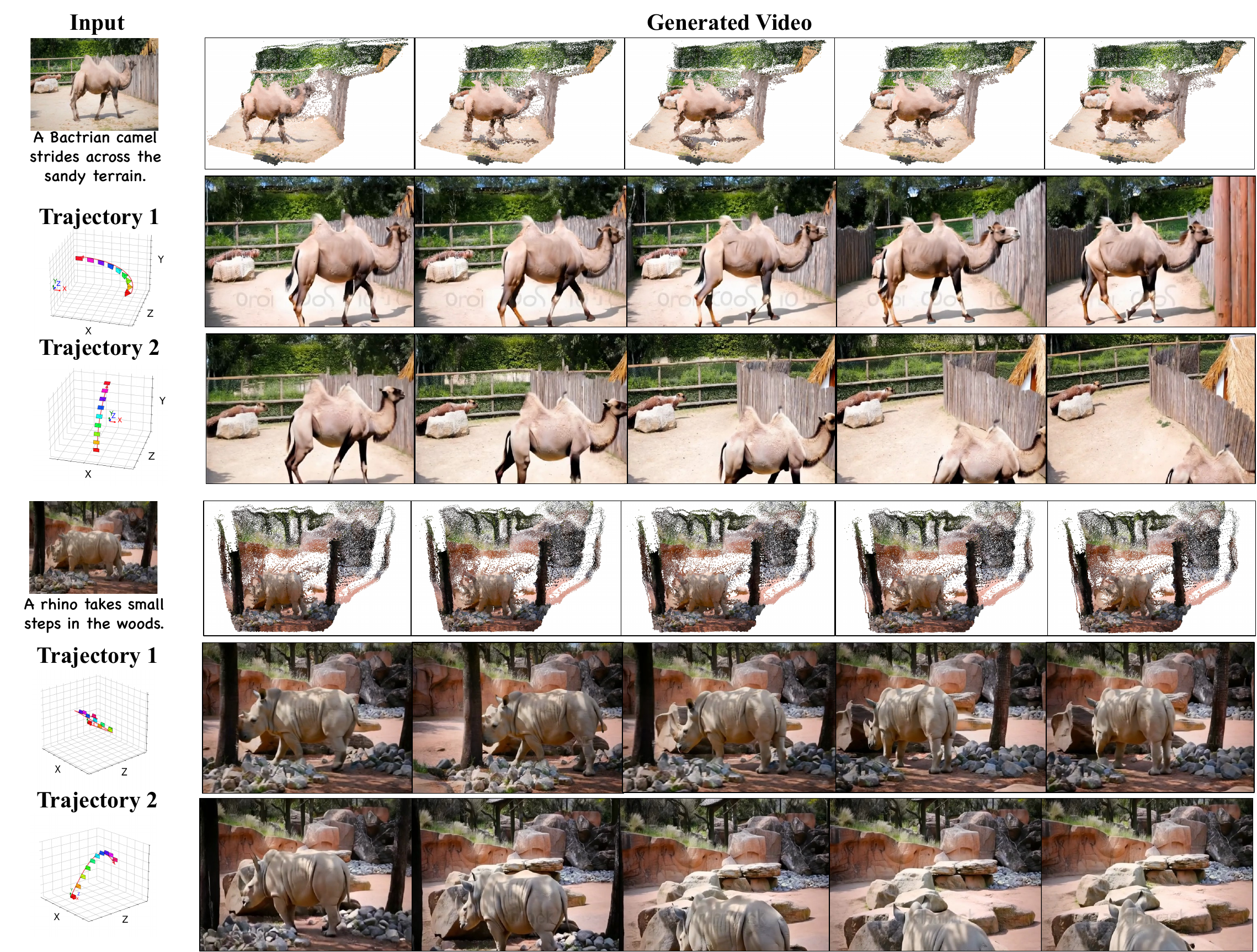}
\vspace{-10pt}
\caption{\textbf{Qualitative results of our model.} 
We visualize the generated results for two more samples. 
The first row shows the 4D point cloud generated by our 4D-STraG. The second, third rows show the videos rendered by our 4D-ViSM under three distinct, user-defined camera trajectories. }
\label{fig:demo}
\vspace{-15pt}
\end{figure*}

\subsection{Qualitative Results of MoGe4D}

\begin{figure*}[t]
\centering
\includegraphics[width=\textwidth]{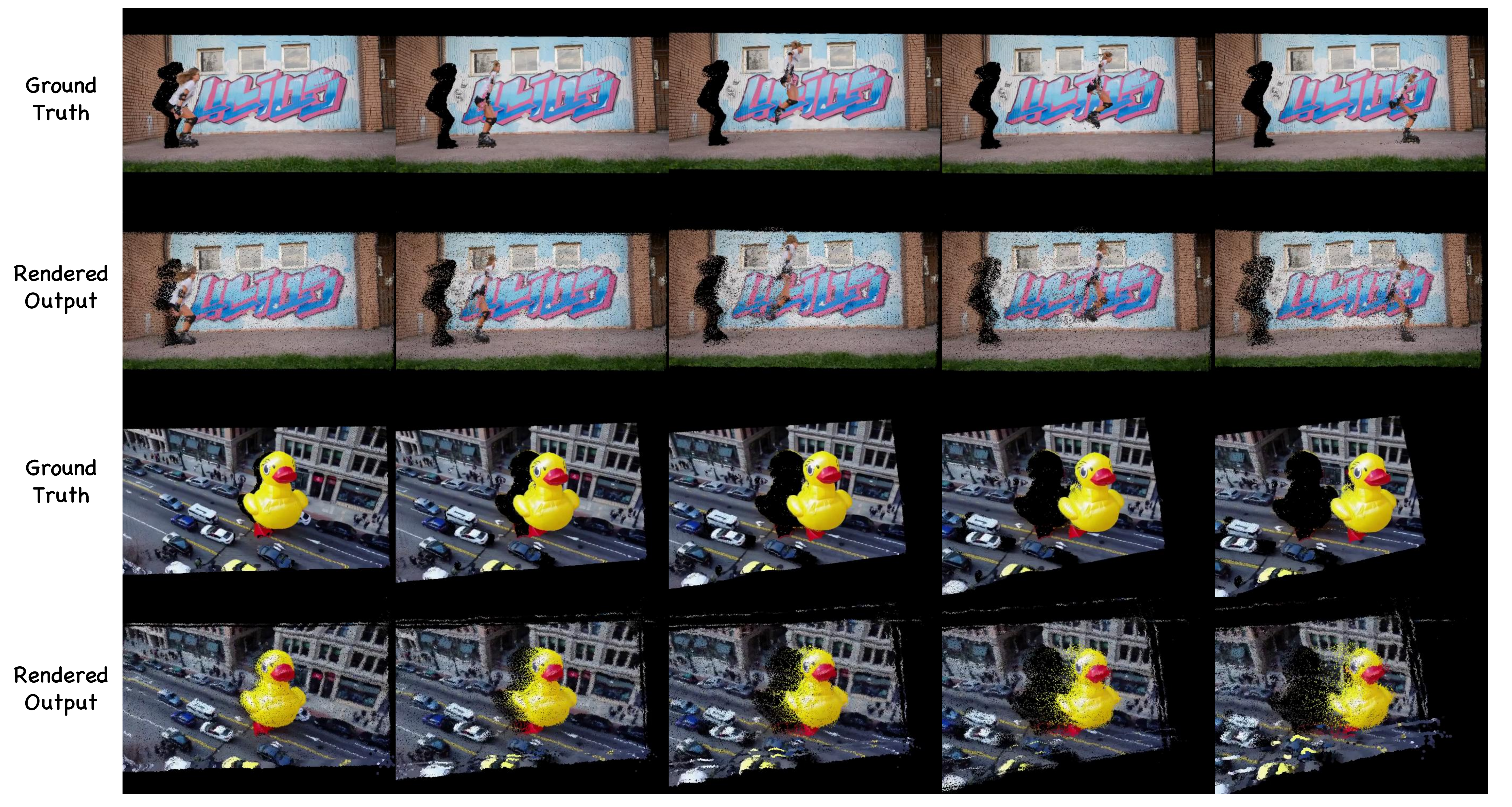}
\vspace{-15pt}
\caption{\textbf{Qualitative results of Motion-Sensitive VAE.}  The figure presents two examples. For each, the top row shows pseudo-GT from the original camera view, while the bottom is the corresponding VAE rendering at the same pose. The close visual alignment demonstrates our VAE's high-fidelity reconstruction capability.
}
\vspace{-5pt}
\label{fig:vae}
\end{figure*}

Figure~\ref{fig:demo} presents two samples of the visualizations for MoGe4D. They are evaluated under distinct user-defined camera trajectories: top-down viewing, forward-backward movement, right-hand circling, and left-hand circling. These trajectories are designed to provide comprehensive multi-perspective observations. The rendered videos demonstrate that our approach consistently produces temporally coherent motion and maintains high-fidelity visual details across all viewpoints. The high-quality 4D point clouds produced by 4D-STraG provide a robust geometric foundation for rendering. The results further demonstrate the strong capability of our 4D-ViSM renderer in synthesizing consistent, high-fidelity videos under diverse camera paths, confirming its flexibility and generalization.

\begin{figure}[t]
  \centering
  \includegraphics[width=\linewidth]{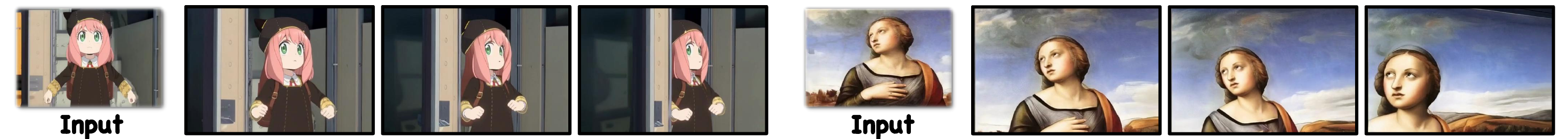}
  \vspace{-10pt}
  \caption{Examples showing generalization to OOD styles.}
   \label{fig:OOD}
   \vspace{-15pt}
\end{figure}

\subsection{Qualitative Results of Motion-Sensitive VAE}

Figure~\ref{fig:vae} shows the reconstruction results using the fine-tuned VAE on a subset of samples during inference. The reconstructed points are rendered from the original camera viewpoint, demonstrating that they effectively preserve the structural and motion consistency with the input. More specifically, the model exhibits robust representation learning capabilities, enabling effective extraction of latent space representations that accurately capture both geometric and dynamic properties of the observed scenes.

\subsection{OOD Generalization}

Figure ~\ref{fig:OOD} demonstrates that \ours generalizes effectively to out-of-distribution stylized inputs, including cartoons and paintings, provided the underlying geometric structure remains sufficiently discernible. These findings suggest that our framework's generalization stems from its geometry-grounded representation, while robustness to photometric extremes and unstructured scenes remains an open challenge for future work.

\begin{figure*}[t!]
\centering
\includegraphics[width=0.95\textwidth]{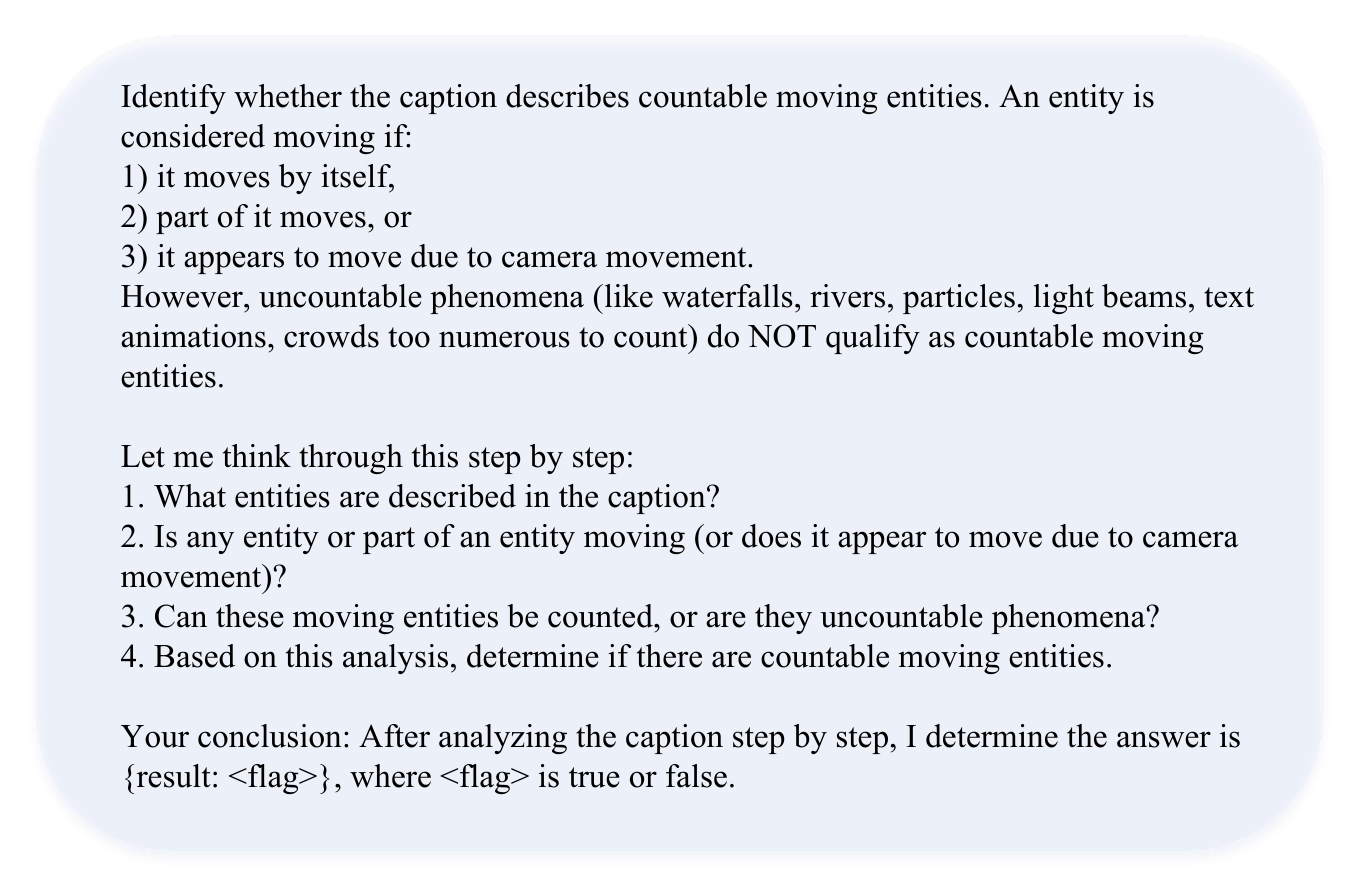}
\vspace{-10pt}
\caption{\textbf{Our input prompt} used to query the model for identifying and counting self-initiated and articulated motion.
}
\label{fig:appendix_prompt}
\vspace{-10pt}
\end{figure*}

\section{Dataset Curation}
\label{sec:d}
\subsection{Video Filtering and Annotation}

We initiated our process with approximately 200,000 video candidates from the WebVid-10M~\cite{webvid} dataset. To automate the selection of content suitable for 4D dynamic scene modeling, we implemented a two-stage pipeline. First, we employed the multimodal large language model CogVLM2~\cite{cogvlm2} to generate a detailed English caption for each video. Subsequently, these captions were fed into the DeepSeek-V3~\cite{deepseekv3} model, which used a carefully designed prompt to assess content suitability. The core evaluation criteria were: (1) the presence of one or more clearly countable entities, and (2) the exhibition of self-initiated, non-rigid, or articulated motion, as opposed to random movements driven by external forces (e.g., wind, water) or dominant camera motion. This process effectively filtered out videos featuring unstructured dynamics, such as crowd movements, water ripples, or swaying foliage. Our prompt is shown in Figure \ref{fig:appendix_prompt}.

\subsection{4D Trajectory Quality Control}

After extracting the raw 4D trajectories, we applied a strict quality filtering process to eliminate samples compromised by depth estimation failures or extreme motion. The specific criteria were as follows: (1) We removed samples where a significant portion of point cloud trajectories contained invalid or anomalous depth values (e.g., near-infinite or zero) at any timestep. (2) We discarded samples exhibiting an excessively large standard deviation in scene depth, which indicates potential errors in the global depth estimation. (3) We performed a scale consistency check. Since uniformly scaling a point cloud should yield an identical rendering from the original camera perspective, we removed samples where this transformation resulted in significant visual changes, flagging them as geometrically inconsistent. This comprehensive filtering pipeline yielded the final 60,000 high-quality samples that constitute the TrajScene-60K dataset.

\begin{table*}[t]
\centering
\caption{\textbf{Comprehensive comparison of 3D scene understanding datasets.} Our TrajScene-60K dataset significantly surpasses existing benchmarks in scale, providing orders of magnitude more frames and 3D point annotations, while offering dense tracking and depth information across diverse real-world indoor and outdoor environments.}
\vspace{-3mm}
\resizebox{\textwidth}{!}{
\begin{tabular}{lcccccccc}
\toprule
\textbf{Dataset} & \textbf{Year} & \textbf{Frames} & \textbf{Resolution} & \textbf{3D Points} & \textbf{Dense Track} & \textbf{Depth}& \textbf{Caption} & \textbf{Description} \\
\midrule
FlyingThings3D~\cite{flyingthing3D} & CVPR'15 & $2.6\times10^4$ & $960\times540$ & $3\times10^8$ & \checkmark & \checkmark &$\times$ & Synthetic, object-level \\
ScanNet~\cite{scannet} & CVPR'17 & $2.5\times10^6$ & $1296\times968$ & $0$ & $\times$ & \checkmark &$\times$& Real, indoor scenes \\
Kubric~\cite{greff2022kubric} & CVPR'22 & Flexible & $256\times256$ & Flexible & \checkmark & $\times$ &$\times$& Synthetic, object-level \\
TAPVid~\cite{tapvid} & NeurIPS'22 & $3\times10^5$ & Multiple & $0$ & \checkmark & $\times$ &$\times$& Real + synthetic videos \\
TAPVid-3D~\cite{tapvid3d} & NeurIPS'24 & $5\times10^5$ & Multiple & $1\times10^6$ & $\times$ & \checkmark &$\times$& Real-world scenes \\
\midrule
TrajScene-60K & - & $3\times10^6$ & $596\times336$ & $1.2\times10^{10}$ & \checkmark & \checkmark & \checkmark& Real, indoor/outdoor \\
\bottomrule
\end{tabular}%
}
\vspace{-5mm}
\label{tab:dataset_comparison}
\end{table*}

\subsection{Comparison with Previous Datasets}

Our TrajScene-60K dataset fills a critical gap in the landscape of resources for 4D dynamic scene understanding and generation. As summarized in Table~\ref{tab:dataset_comparison}, existing datasets are often limited in scale, realism, or the richness of provided annotations. For instance, while synthetic datasets like FlyingThings3D~\cite{flyingthing3D} and Kubric~\cite{greff2022kubric} offer precise ground-truth motion and geometry, their domain gap with real-world imagery hinders generalization. Real-world datasets like ScanNet~\cite{scannet} provide dense 3D reconstructions but are static and lack temporal dynamics. Recent video tracking benchmarks like TAPVid~\cite{tapvid} and TAPVid-3D~\cite{tapvid3d} offer real-world point trajectories but are orders of magnitude smaller in scale and do not provide semantic descriptions crucial for conditional generation.

TrajScene-60K distinguishes itself by combining large-scale, real-world video content with dense, occlusion-aware 4D trajectories and high-quality text captions. It provides over \textbf{3 million frames} and approximately \textbf{12 billion 3D point annotations}, significantly surpassing predecessors in volume. Crucially, we provide dense 4D tracking, per-frame depth, and language descriptions for dynamic scenes across diverse indoor and outdoor environments. This unique combination of scale, realism, and multi-modal annotation makes it an enabling resource for training and evaluating complex 4D scene generation or tracking models, facilitating a more holistic understanding of dynamic 3D worlds.

\subsection{Dataset Bias and Release}

TrajScene-60K is built from WebVid-10M~\cite{webvid} using LLM/VLM-based caption filtering, and therefore inherits both the demographic and content biases of the source videos and the filtering models. For example, certain object categories, body types, or geographic regions may be under-represented, and the motion statistics may be skewed towards popular online content. While our work focuses on the technical aspects of 4D scene generation, we acknowledge these biases and will release the dataset and filtering prompts to facilitate transparency and future audits. We also encourage downstream users to carefully consider fairness and representational biases when deploying models trained on TrajScene-60K.

\section{Baseline Introduction}
\label{sec:e}
We compare our method with the following leading methods focusing on 4D generation from a single image. 

\noindent\textbf{4Real}:~\cite{4real} generates photorealistic 4D scenes using video diffusion models trained on real data. It first creates a reference video, then learns its 3D canonical representation and temporal deformations, avoiding synthetic 3D priors.

\noindent\textbf{GenXD}:~\cite{genXD} introduces a real-world 4D dataset and multi-view temporal modules to disentangle camera and object motion, enabling joint learning from 3D/4D data for arbitrary scene generation.

\noindent\textbf{Gen3C}:~\cite{gen3c} guides video generation via a 3D cache from predicted point clouds, ensuring precise camera control and 3D consistency by focusing generation on unobserved regions.

\noindent\textbf{DimensionX}:~\cite{dimensionX} reconstructs 3D/4D scenes from a single image by decoupling spatio-temporal factors in a controllable video diffusion framework, allowing precise manipulation of structure and dynamics.

\noindent\textbf{Free4D}:~\cite{free4d} proposes a tuning-free framework that distills foundation models to generate consistent 4D scenes from one image, enabling strong generalization without expensive training.

Among these, 4Real, GenXD, DimensionX, and Free4D follow a \textit{generate-then-reconstruct} pipeline, while Gen3C adopts a \textit{reconstruct-then-generate} strategy.

\section{ Implementation Details}
\label{sec:f}
\subsection{Motion-Sensitive VAE Architecture}

Before training the DiT in the 4D-STraG module, we finetune a specialized, motion-sensitive VAE to effectively adapt our generative backbone for trajectory synthesis. This VAE is designed to process and reconstruct trajectory information encoded as RGB-like motion maps. Inspired by Geo4D~\cite{geo4d}, the architecture of both the VAE encoder and decoder is intentionally kept shallow to preserve fine-grained motion details. The Trajectory Encoder and Decoder that process these motion maps are constructed as shallow ResNets~\cite{resnet}. This minimalist design ensures that the VAE learns a compact and efficient latent space for motion patterns without aggressively downsampling the spatial features, which is critical for the precise reconstruction of trajectories by the Trajectory Decoder. When finetuning Motion-Sensitive VAE, we trained the Trajectory Encoder, Trajectory Decoder, and the VAE Decoder, while freezing the VAE Encoder. This approach allows the model to fully adapt to the motion input while preserving the integrity of the pre-trained visual representations.

\subsection{Details of 4D-ViSM Architecture}
Our 4D-ViSM model adapts a pre-trained video Diffusion Transformer (DiT) for dynamic video inpainting, specifically to fill holes in novel-view videos rendered from our 4D representation. During fine-tuning, the model is conditioned on the incomplete video. For each step, the rendered video with holes is encoded into a latent representation $z_{\text{rendered}}$, and its binary occlusion mask is downsampled to $m_{\text{latent}}$. The denoising network's input is a channel-wise concatenation of the noisy latent $z_t$, $z_{\text{rendered}}$, and $m_{\text{latent}}$, formulated as 
$$z = \text{concat}([z_t, z_{\text{rendered}}, m_{\text{latent}}]).$$
This explicitly provides the model with the known visual context and the missing regions, enabling coherent video completion.

\subsection{Inference Pipeline}

Given a single image and a text prompt, our model first acquires geometric information by estimating depth through the 4D-STraG module. Specifically, we utilize the UniDepthv2~\cite{unidepthv2} model to infer depth information, ensuring consistency with the estimation method used for the DELTA tracking model during training. The VAE then encodes the image and depth map into a latent space, while MPM extracts motion features. Subsequently, DiT generates latent representations, which are decoded into relative motion latents. These latents are de-normalized by reversing our depth-guided motion normalization strategy and then fused with the initial point cloud's spatial coordinates to construct a 4D scene representation. Finally, this representation allows for rendering from arbitrary camera poses, and our 4D-ViSM synthesizes a spatio-temporally consistent 4D video aligned with the desired camera trajectory.

\section{Applications and Future Work}
\label{sec:g}

Our image-to-4D generation framework opens up numerous possibilities across various domains. It can serve as a powerful tool for digital content creation, enabling artists and designers to bring static images to life as dynamic 3D assets for films, games, and advertising. Furthermore, this technology holds significant potential for populating virtual and augmented reality (VR/AR) environments with dynamic objects, creating more immersive and interactive digital experiences. Looking ahead, we identify two primary directions for future research. First, we plan to explore more deeply unified architectures that further entangle motion and appearance generation, potentially within a single, non-autoregressive model, to achieve even greater spatio-temporal consistency. Second, a critical challenge in this field is the lack of standardized evaluation protocols. We aim to develop novel metrics specifically designed to quantify the 4D consistency of generative models, providing a more reliable and automated way to measure geometric and textural stability over time.

\section{Prompt of VLM-based 4D Evaluation}
\label{sec:prompt}

We present the comprehensive prompt for VLM-based 4D evaluation. The detailed prompt is as follows:

\begin{lstlisting}
You are an expert Computer Vision researcher and evaluator specializing in 4D scene generation and dynamic 3D reconstruction. Your task is to critically evaluate a video sequence generated from a single static image. The video depicts a 3D scene that is both moving (dynamic) and being viewed from a moving camera.

You will be provided with a video sequence (or a series of frames) representing a generated 4D scene. You must assess the quality of this generation based on five specific metrics related to spatiotemporal consistency and realism.

Please analyze the visual input and provide a score from **1 to 5** for each of the following metrics. A score of 1 represents "Poor/Failure" and 5 represents "Excellent/Flawless."

**1. 3D Geometric Consistency**
* Does the scene maintain a rigid and consistent 3D structure (for static parts) and plausible deformation (for dynamic parts) as the camera moves?
* Look for: Distortions in perspective (parallax errors), objects "flattening" or warping unnaturally when the viewing angle changes, or parts of the geometry drifting apart.
* 5 (Excellent): Perfect perspective; depth cues are consistent across all frames; no geometric warping.
* 1 (Poor): Severe perspective distortion; the scene looks like a warping 2D image rather than a 3D space.

**2. Temporal Texture Stability (Appearance Consistency)**
* Do the surface textures and colors remain consistent over time and across viewpoints?
* Look for: "Texture swimming" (noise/patterns moving independently of the object), flickering colors, or inconsistent lighting on the same surface across frames.
* 5 (Excellent): Textures are stuck firmly to the geometry; lighting is coherent.
* 1 (Poor): Severe flickering, color shifting, or textures that slide over the object surface.

**3. Subject Identity Preservation**
* Does the main subject (the object or character) maintain its semantic identity and structural integrity compared to the implied initial state?
* Look for: The object morphing into something else, losing limbs, or faces becoming unrecognizable during motion.
* 5 (Excellent): The subject is clearly recognizable and maintains high fidelity throughout the video.
* 1 (Poor): The subject loses its shape or identity completely by the end of the sequence.

**4. Motion-Geometry Coupling (Physics Plausibility)**
* Does the motion make sense given the 3D geometry?
* Look for: "Sliding" artifacts (feet moving but not propelling the body), parts of the object passing through each other (clipping), or background elements moving along with the foreground object (disocclusion errors).
* 5 (Excellent): Motion feels physically grounded in the 3D space; correct occlusion/disocclusion.
* 1 (Poor): Floating objects, severe clipping, or motion that ignores the 3D structure.

**5. Background Stability**
* As the camera moves, does the background behave like a static environment?
* Look for: The background moving in sync with the foreground subject, background jitter, or the "canvas" tearing at the edges.
* 5 (Excellent): Background acts as a stable, static 3D environment (or moves naturally if it's a dynamic background).
* 1 (Poor): Background distorts significantly or moves attached to the foreground.

Please provide your response in the following JSON format. Do not output markdown formatting around the JSON.

{
  "reasoning": "A brief, concise explanation of your observations regarding artifacts, consistency, and motion quality.",
  "scores": {
    "3D_Geometric_Consistency": <score>,
    "Temporal_Texture_Stability": <score>,
    "Subject_Identity_Preservation": <score>,
    "Motion_Geometry_Coupling": <score>,
    "Background_Stability": <score>
  }
}
\end{lstlisting}